\documentclass[runningheads]{llncs}
\usepackage{graphicx}
\usepackage{amsmath,amssymb} 
\usepackage{color}

\usepackage{comment}
\usepackage{xcolor,colortbl}

\usepackage{xspace}
\usepackage{mathtools}
\usepackage{algorithm}
\usepackage[noend]{algpseudocode}

\def\R{\mathbb{R}}
\newcommand{\dx}{\mathrm{d}x}

\DeclareMathOperator*{\argmin}{arg\,min}

\newcommand{\Tab}{Tab.\@\xspace}

\newcommand{\Eqs}{Eqs.\@\xspace}
\newcommand{\Fig}{Fig.\@\xspace}

\newcommand{\Sec}{Sec.\@\xspace}
\newcommand{\trans}{^{\!\mathsf{T}}}

\newcommand{\wrt}{w.r.t.\@\xspace}

\newcommand{\cf}{c.f.\@\xspace}

\newcommand{\eg}{e.g.\@\xspace}

\newcommand{\ie}{i.e.\@\xspace}

\newcommand{\wlogg}{wlog.\@\xspace}

\definecolor{myGreen}{rgb}{0.2,0.6,0.}
\definecolor{myCyan}{rgb}{0.0,0.4,0.9}
\newcommand{\myparagraph}[1]{\vspace{0.5em}\noindent\textbf{#1}}

\newcommand{\invisible}[1]{}
\newcommand{\RNum}[1]{{\bf (\lowercase\expandafter{\romannumeral #1\relax})}}

\newcommand{\vecII}[2]{{\left( \begin{array}{c} #1 \\ #2 \end{array}\right)}}

\begin{document}

\title{Learning Energy Based Inpainting for Optical Flow\thanks{Supported by the ERC starting grant 640156, ’HOMOVIS’}} 
\titlerunning{Learning Energy Based Inpainting for Optical Flow}



\author{Christoph Vogel\inst{1}\orcidID{0000-0002-5960-1375} \and
Patrick Kn\"{o}belreiter\inst{1}\orcidID{0000-0002-2371-014X} 
\and
Thomas Pock\inst{1,2}\orcidID{0000-0001-6120-1058}}
%

\authorrunning{C. Vogel et al.} 

\institute{Graz University of Technology, Graz, Austria. 
\email{\{firstname.lastname\}@icg.tugraz.at} 
\and Austrian Institute of Technology, Vienna, Austria}
%

\maketitle

\begin{abstract}
%
Modern optical flow methods are often composed of a cascade of 
many independent steps 
or formulated as a black box neural network that is hard 
to interpret and analyze. 
In this work we seek for a plain, interpretable, but learnable solution. 
We propose a novel inpainting based algorithm that approaches the problem in three steps:
feature selection and matching, selection of supporting points 
and energy based inpainting. 
%
%
%
To facilitate the inference we propose an \emph{optimization layer} 
that allows to backpropagate through 10K iterations of a first-order 
method without any numerical or memory problems. 
Compared to recent state-of-the-art networks, 
our modular CNN is very lightweight 
and competitive with other, more involved, inpainting based methods. 
	\end{abstract}
	
	\section{Introduction \& Related Work}
	\label{sec:intro}
	The computation of optical flow, the apparent 2D motion field 
between two consecutive frames of a temporal image sequence, 
is one of the most investigated problems in computer vision. 
Optical flow possesses a vast number of applications, among others, 
video processing for action detection or activity recognition, 
autonomous driving~\cite{Menze2015CVPR} or medical imaging. 
Similarly, there are multiple methods, to compute the optical flow field.
Energy based techniques that were popular in the past, are now 
outperformed and replaced by approaches that solely rely
on convolutional neural networks (CNNs)~\cite{DFIB15,IMSKDB17}. 
Despite their great performance on benchmark data sets, these 
networks often lack interpretability -- often it remains unclear, 
apart from input and loss function, 
how the network internally models the flow problem. 
%
%
In this work we seek to combine both ideas. 
We propose to utilize an energy based optimization problem suitable for the 
computation of optical flow and let the network learn the input to the energy. 
%
We then minimize the energy to produce the flow field by 
unrolling the iterations of a first-order algorithm. 
Our energy is convex but non-smooth and hence demands for a large
number of iterations.
To address the memory and numerical problems that occur when running 
backpropagation for more than 10K iterations on the GPU, we 
propose an \emph{optimization layer}, that handles these problem in an 
efficient GPU implementation using \emph{checkpointing} 
\cite{Griewank2000,ChenXZG16} and 
buffering of intermediate solutions in double precision.
%
%
Our energy minimizing network layer is related to ideas proposed 
in \cite{BarronPoole2016}. Compared to \cite{BarronPoole2016}, 
we solve a non-linear and non-smooth problem, but similarly 
facilitate to backpropagate through the minimization process. 
However, our experiments indicate that the increased robustness provided 
by the non-smooth formulation is beneficial for our optical flow problem. 
A combination of CNNs and energy optimization 
was also proposed by \cite{crfasrnn_iccv2015,riegler16dsr,VogelP17}. 
Here, we seek to run our optimization until near convergence. 
Our optimization layer allows us to unroll 4 orders of magnitude
more iterations than \cite{crfasrnn_iccv2015,riegler16dsr}. 

For our model we consider a class of algorithms 
that treat the computation of optical flow as a form of 
\emph{inpainting} or sparse-to-dense interpolation problem. 
%
%
Possibly the most prominent representative of these methods is 
\cite{RevaudWHS15} that tackles the problem in many different steps, 
including sparse matching \cite{weinzaepfel13}, edge detection \cite{Dollar2013}, 
computing super-pixels \cite{Achanta2012}, variational refinement \cite{BroxBPW04}, 
and various post-processing steps. 
Here, we simplify the inpainting process to its core, 
feature generation to build a cost volume for image matching, 
selection of supporting matches and inpainting via energy minimization. 
We tackle the problem via deep learning and propose a network 
structure that still delivers interpretable intermediate results, 
and allows for training in end-to-end fashion. 
In contrast to other inpainting based optical flow methods, 
we start our process from dense matching. 
Consequently, we are not committed to a pre-selection of, 
possibly incomplete or unmatchable interest points \cite{weinzaepfel13,ZweigW17},
but can select the supporting pixels \emph{after} matching. 
Compared to nearest neighbor field methods~\cite{Hu_2016_CVPR,global-patch-collider}, 
we make use of a complete cost-volume and avoid a coarse-to-fine scheme or hashing. 
To that end, we make use of a recent result 
\cite{MundaSKP17} that allows for a low memory 
footprint of the cost volume. 
In contrast to network based solutions to inpainting \cite{ZweigW17}, 
we maintain the interpretability of an energy based framework.


The core idea of our algorithm is closely related to diffusion 
based inpainting for image compression \cite{Galic2008}. 
%
Here, the compression ratio is mainly dependent on the selection of 
\emph{supporting points} from which the image is then inpainted. 
%
%
Given the image to be compressed, this selection process can be formulated as a 
bi-level optimization problem \cite{ChenRP14a_corr,Peter2015}. 
Unfortunately, we have no knowledge of the ground-truth reconstruction 
and have to select the supporting points from a large number of 
possible matches per pixel. 
Here, our method estimates the confidence of the different matches per pixel. 
In the context of stereo matching confidence estimation 
is a well studied problem~\cite{Hu2012}. 
%
Lately also CNN based solutions have been proposed~\cite{Agresti_2017_ICCV,Poggi_2017_ICCV}. 
Compared to the general problem, our solution is directly task related. 
We are not interested in providing a confidence for each pixel, 
but also have to consider the relevance of a pixel to our inpainting task. 
Our selection process has to balance the added information content and 
robustness of the match on a per pixel basis. 
%
This is addressed by learning confidence estimation and inpainting 
in end-to-end fashion. 



In this paper, we propose a novel, 'from scratch' algorithm for
inpainting optical flow.
We reduce the process to its core: feature computation and matching, 
selection of supporting pixels and energy based inpainting. 
%
Compared to recent state-of-the-art networks, our CNN is lightweight 
with only 450K parameters. 
We introduce a novel \emph{quad-fitting} layer that can learn 
features for sub-pixel accurate matching, 
while still allowing for large displacements 
and finally propose an \emph{optimization layer} to solve the energy 
minimization problem.
Here, we show that a tailored GPU implementation can lead to memory 
and numerical efficiency and facilitate backpropagation over 
more than 10K iterations. 

	\section{Method}
	\label{sec:method}





In this work we consider the estimation of optical flow, 
the 2D motion field that describes the movement of pixels 
between two consecutive frames $I^0, I^1\in\Omega\subset\mathbb{R}^2$ 
of an image sequence defined over the domain $\Omega$. 
%
Formally, we define the optical flow field as
$\mathbf{u}:=(u_0,u_1)\trans$, consisting of a vertical 
and a horizontal component given by the functional
$u_i:\Omega\rightarrow\mathbb{R}, i\in\{0,1\}$. 
Here, we let super-indices encode the time step and sub-indices 
the motion direction. 
We formulate the task of estimating such motion field 
as a classical inpainting/denoising problem: 
%
%
%
\begin{equation}\label{eq:func_tv}
  \min_{u_i} \mathcal{R}_1(u_i) + \int_\Omega
  c(x)|u_i(x)-\hat{u_i}(x)| \dx, \; \textrm{ with }\,
  \mathcal{R}_1(u_i) := \int_\Omega |W^\frac{1}{2} \nabla
  u_i|_\delta \dx,
\end{equation}
which corresponds to weighted Total Variation~\cite{Rudin1992}
regularization with a robust weighted $\ell_1$ data
fidelity term, known to lead to piecewise constant solutions.
Here, $|\cdot|_\delta$ denotes the Huber-norm $|\cdot|_\delta
:= \tfrac12 |\cdot|_2^2+ \tfrac12 \delta^2$, if $|\cdot|_2\leq\delta$
and $\delta |\cdot|_2$ else.
Likewise, we also consider a variant corresponding to weighted Total
Generalized Variation~\cite{Bredies2010} of second order, were we
replace the regularization term by:
\begin{equation}\label{eq:func_tgv}
  \mathcal{R}_2(u_i) = \min_{w_{i}=(w_{i,0},w_{i,1}\!)\trans} \int_\Omega\!\!\!
  |W^\frac{1}{2} \nabla u_i \!-\! w_{i}|_\delta + \beta ( |\nabla
  w_{i,0} |_\delta + |\nabla w_{i,1} |_\delta )\dx,
\end{equation} 
where $w_{i,j}\in\Omega\rightarrow\mathbb{R}, i,j\in\{0,1\}$ 
represent auxiliary variables. 
In contrast to \eqref{eq:func_tv}, \eqref{eq:func_tgv} prefers a
piecewise affine solution for the flow components $u_i$. 
%
Both cases require a good guess
for the initial flow $\hat{\mathbf{u}}\!:=\!(\hat{u}_0,\hat{u}_1)\trans$, 
the \emph{diffusion tensor} $W:=\textrm{diag}(\omega_0,\omega_1)$, 
\cf\cite{Werlberger2009a} 
and the \emph{confidence score} $c\!\in\![0,1]$ that locally ties 
the solution $\mathbf{u}$ to the initial estimate $\mathbf{\hat{u}}$. 
%
%
To compute $\mathbf{u}$ we use a CNN that is split into different 
parts to deliver the inputs for our optimization stage solving 
\eqref{eq:func_tv} or \eqref{eq:func_tgv}. 
%
%
In particular, we perform dense pixel-wise matching using network generated 
features, refine the matches by locally fitting a quadratic to the cost 
and employ the $\argmin$ cost solution as initial estimate. 
Diffusion Tensor $W$ and confidence $c$ are provided by 2 sub-networks. 
All these CNNs are rather small, with around 150k parameters each. 
To solve our optimization problem we employ a special 
custom layer that allows to accurately minimize convex, non-smooth problems 
in the form of \eqref{eq:func_tv} or \eqref{eq:func_tgv} and provides a simple, 
memory efficient and accurate way to backpropagate through them and, thus, 
learn its input parameters. 
%
%
%
\begin{figure}[t]
  \begin{center}
     \includegraphics[width=0.9297\linewidth]{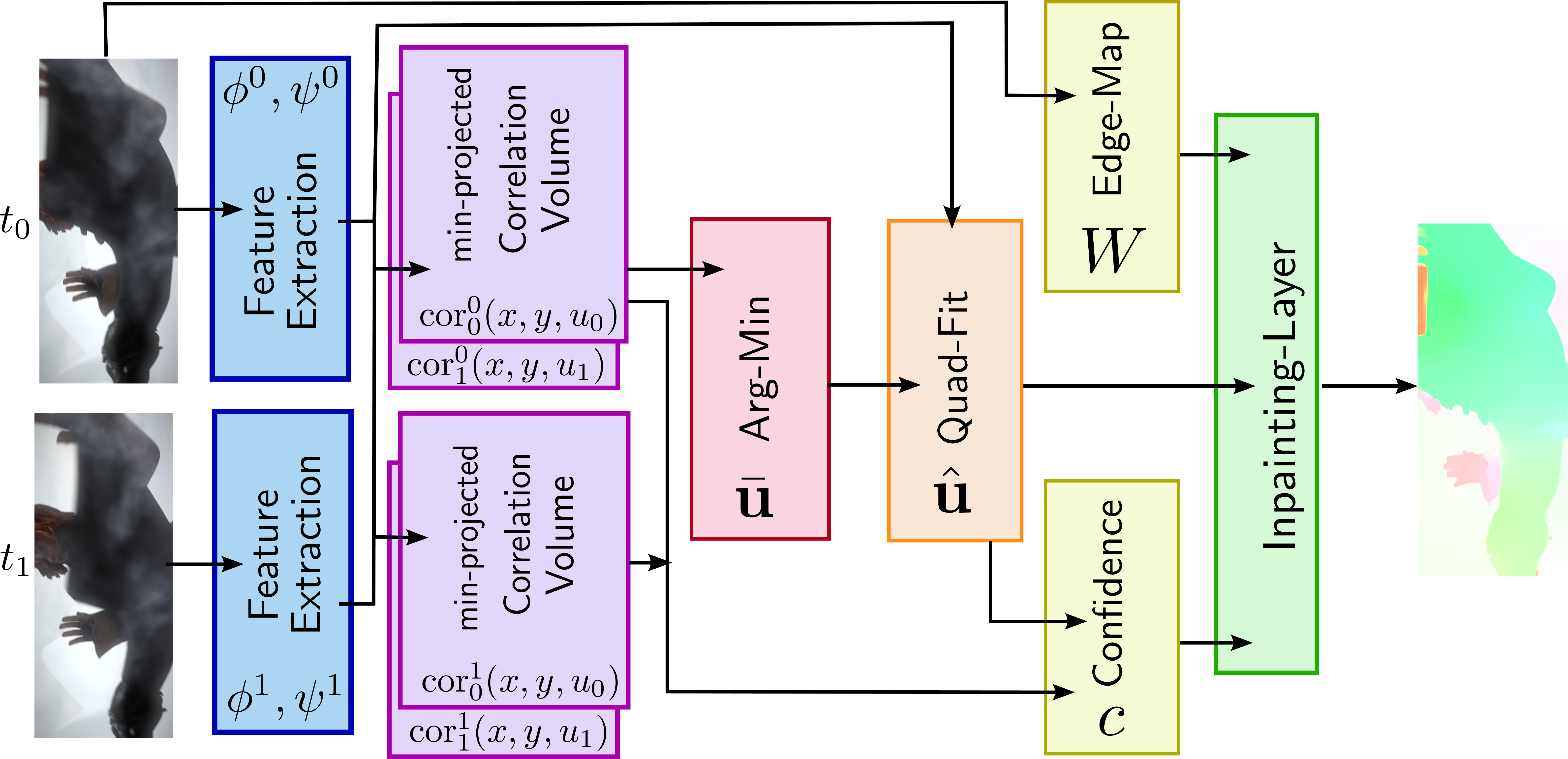} 
  \end{center}
     \caption{
      The overview of our pipeline. Feature vectors are generated from the input images. 
      Per view, in forward and backward direction, two (see \Sec\ref{sec:FG}) correlation 
      volumes are built. 
      A preliminary solution $\hat{\mathbf{u}}$ is obtained by refining the flow at
      minimal (negative) correlation $\bar{\mathbf{u}}$ 
      in the \emph{quad-fitting layer} (see \Sec\ref{sec:quad-fit}). 
      Afterwards, a confidence $c$ is computed for each flow vector obtained from the fitting.  
      In parallel, a diffusion tensor or edge-map $W$ is generated (\Sec\ref{sec:diff}).
      Confidence, tensor and preliminary solution $\hat{\mathbf{u}}$ are fed into 
      our optimization layer 
      (\Sec\ref{sec:opt}) that solves problem \eqref{eq:func_tv} or \eqref{eq:func_tgv}. 
     }
  \label{fig:overview}
\end{figure}

An overview of our method and the progression of our framework is 
provided in \Fig\ref{fig:overview}. 
The input images pass the feature extraction stage from which the cost 
volumes are constructed. 
We employ the method of \cite{MundaSKP17} and generate 2 3D volumes from 
one complete 4D cost volume (\cf\Sec~\ref{sec:FG}). 
This allows to train the feature generation step on whole images 
instead of small patches and also enables us to compute cost volumes 
in forward and backward direction, which are later used for our 
confidence estimates. 
%
To keep the network simple and small 
we do not consider filtering of the cost volumes,
\eg~\cite{Sun2018PWC-Net,Guney2016ACCV,rhemann2011fast}. 
Instead, we directly use the solution with maximal correlation for both 
forward ($\mathbf{\bar{u}}$) and backward direction ($\mathbf{\bar{u}}^\textrm{BW}$). 
%
%
Our \emph{quad-fitting} layer (\Sec\ref{sec:quad-fit}) 
delivers sub-pixel accurate matches, by fitting a quadratic function 
to the local feature cost. 
The computed pixel-wise estimates $\mathbf{\hat{u}}$, along with some extra 
features like the softmax-probabilities are fed into a network to 
compute the confidence scores $c$, \cf \Sec~\ref{sec:confidence}. 
Further, we generate the diffusion tensor, or edge-map, $W$ applying
a network that receives the images as input, described in 
\Sec~\ref{sec:diff}. 
The initial estimates $\mathbf{\hat{u}}$ the confidence scores $c$ and 
the diffusion tensor $W$ are the fed into our optimization layer to 
compute the final solution (\Sec~\ref{sec:opt}).

%
%
Following the flow of our pipeline, we now consider each of its stages in detail.

\def\picsin{0.4447}
\begin{figure}[t]
  \begin{center}
    \includegraphics[width=\picsin\linewidth]{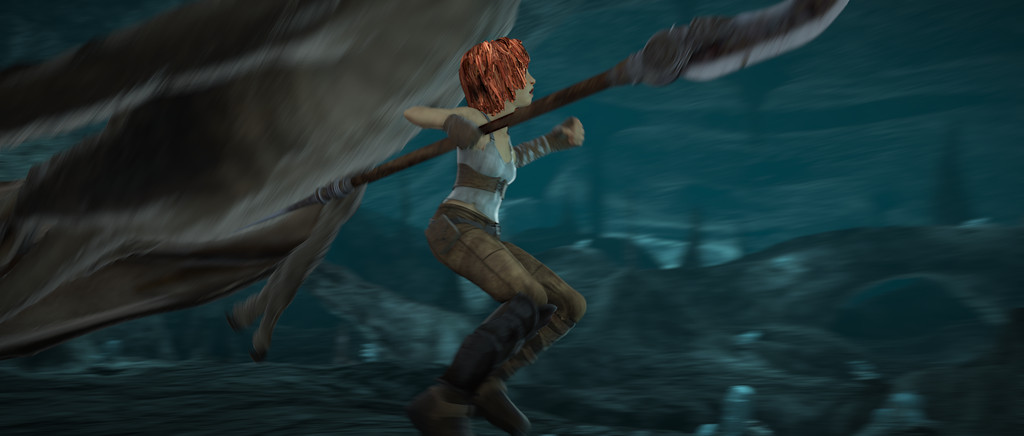}
    \includegraphics[width=\picsin\linewidth]{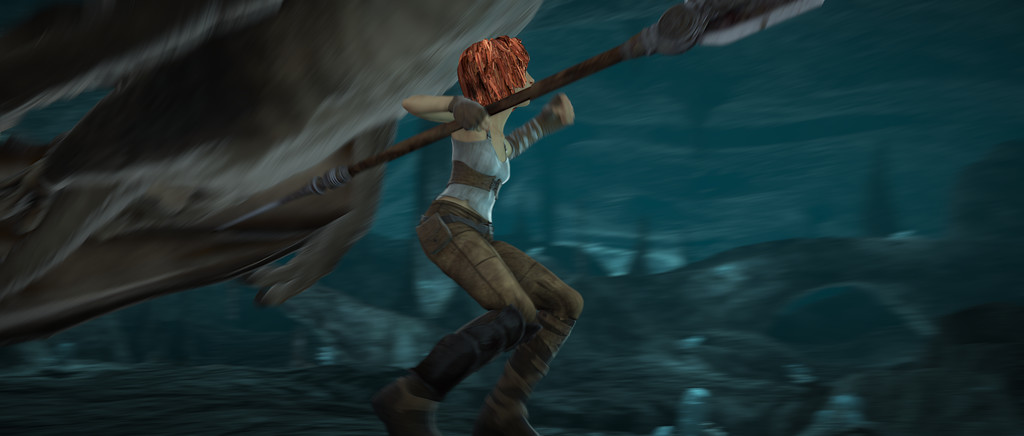}
     \\
     \includegraphics[width=\picsin\linewidth]{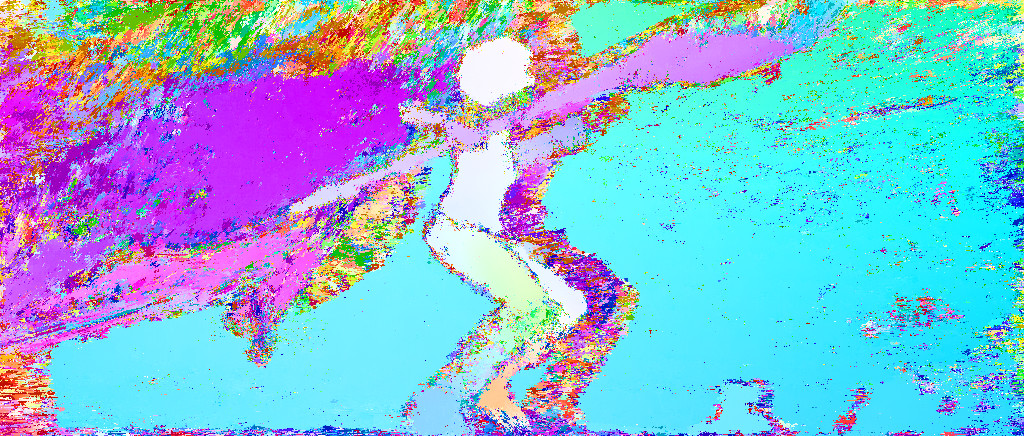}
     \includegraphics[width=\picsin\linewidth]{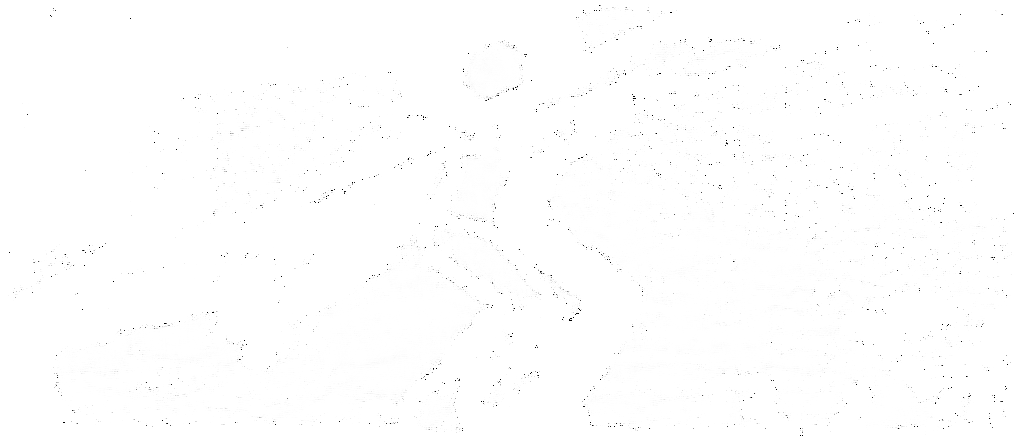}
     \\
     \includegraphics[width=\picsin\linewidth]{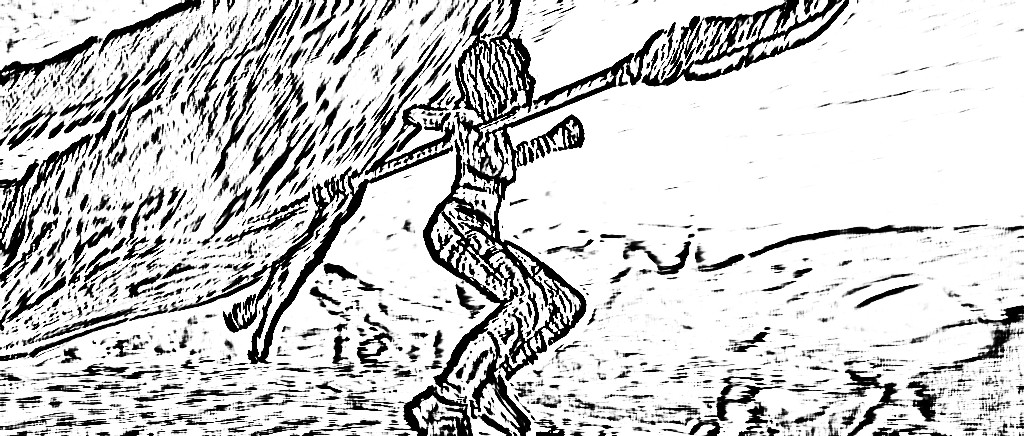}
     \includegraphics[width=\picsin\linewidth]{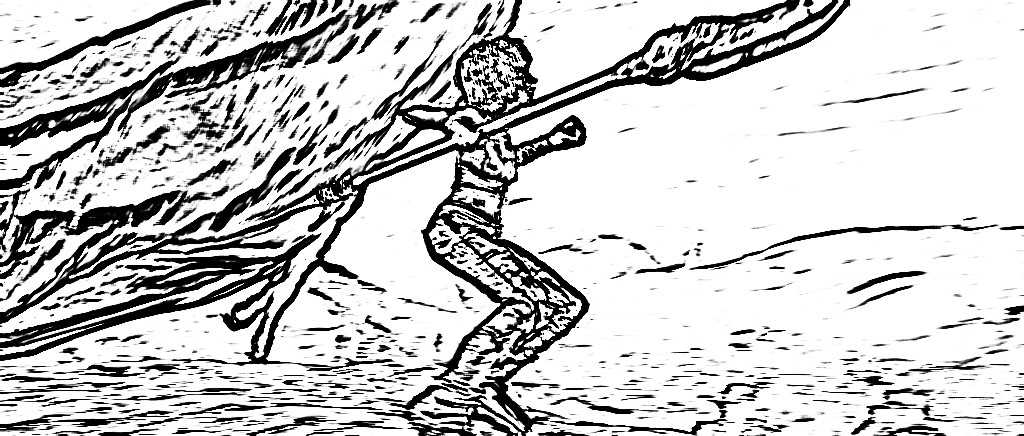}     
     \\
     \includegraphics[width=\picsin\linewidth]{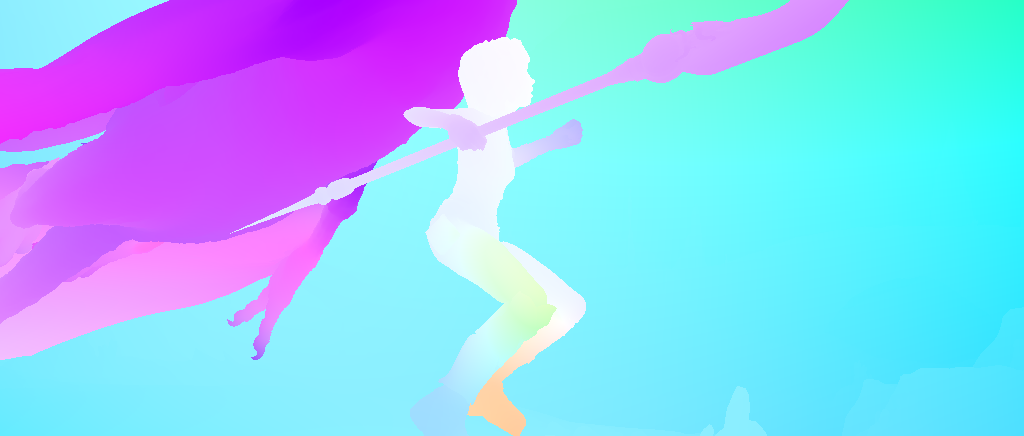}
     \includegraphics[width=\picsin\linewidth]{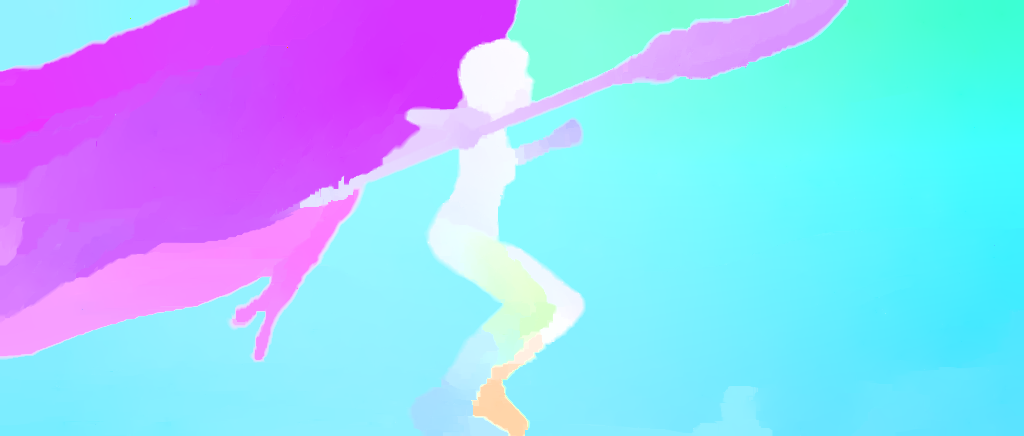}
  \end{center}
     \caption{
      \emph{Top}: Input images at time step 0 (left) and time step 1 (right). 
      \emph{Second row}: Data term $\hat{\mathbf{u}}$ (left) and confidence map $c$ (right), where black means high and white low confidence. 
      \emph{Third row}: Diffusion tensor $W$. Horizontal edges, $W_0$ (left) and vertical edges, $W_1$ (right). Here white means strong and black weak regularization.
     \emph{Bottom}: The ground truth solution (left) and the output of the optimization layer.
     }
  \label{fig:SintelDemo}
\end{figure}
\subsection{Feature Generation}\label{sec:FG}
For feature generation we follow \cite{MundaSKP17} and employ a Siamese 
network consisting of two convolutional branches with shared parameters. 
%
%
In our implementation we utilize a feed forward network comprised of $5$ 
convolutional layers with a filter size of $3\times 3$ and $64$ channels,  
followed by a single 1-D convolution. 
After each stage we apply a $\tanh$ non-linearity. 
Further, we utilize  
dilated filters \cite{YuKoltun2016} of sizes $1,2,4,1,4$ and $1$ 
in the stages. 
In the fourth layer we apply a striding of $2$ and reduce the feature maps 
by a factor of $2$ in each dimension of our domain $\Omega$. 
We retain the feature maps ($\psi^0$ and $\psi^1$) after the third layer 
for our sub-pixel refinement step described in \Sec~\ref{sec:quad-fit}. 
The final feature maps $\phi^0$ and $\phi^1$ of the two frames are fed 
into a correlation layer that generates four 3D cost volumes: 
\begin{equation}\label{eq:corr}
\textrm{cor}^j_i(x,y,u_i):= \min_{u_{1-i}\in H} - \langle \phi^j(x,y), \phi^{1-j}( x+u_0,y+u_1 )\rangle, \quad i,j\in\{0,1\}. 
\end{equation}
%
$H:=\{-d,\ldots,d-1\}$ denotes our range of integer 
displacements, which effectively correspond to a twice as large 
displacement at the original resolution due to the striding. 
Note that we explicitly consider the forward ($\textrm{cor}^0_{i}$) 
and backward direction ($\textrm{cor}^1_i$) and split
the full 4D cost volumes $\textrm{cor}^{j}(x,y,u_0,u_1)$ into 
two directional ones, \cf \eqref{eq:corr}. 
To that end, \cite{MundaSKP17} proposes to use \emph{min-projection} 
to eliminate one of the motion directions via:
$\textrm{cor}^j_{i}(x,y,u_i):= \min_{u_{1-i}\in H} \textrm{cor}^{j}(x,y,u_0,u_1)$.
%
Most prominently this reduces the memory complexity of storing the 
cost volume from quadratic to linear.
Our initial discrete motion estimates can then be found as 
$\bar{u_i}:=\argmin \textrm{cor}^0_i(x,y,u_i), i=0,1$. 
Note, that the $\argmin$ remains untouched by the \emph{min-projection} 
operation applied beforehand on the cost volumes. 
Further, we can define a probability for each pixel 
and possible displacement via the usual \emph{softmax} 
function for the reference view $0$: 
\begin{equation}\label{eq:likelihood}
p_i(x,y,u_i):=\exp(- \textrm{cor}^0_i(x,y,u_i))/\sum_{u_i\in H} \exp( - \textrm{cor}^0_i(x,y,u_i)), \;\textrm{for }i=0,1. 
\end{equation}
Those (pseudo-)likelihoods, 
evaluated at the preliminary motion estimate $\bar{\mathbf{u}}$, 
are used as input for our confidence network and for 
training the data term via a maximum likelihood criterion (\cf\Sec\ref{sec:train}). 
%
%
%
%
%
\subsection{Quad-Fitting}\label{sec:quad-fit}
Our procedure above limits our candidate flow to take only 
even integral values, due to the sampling and additional striding 
imposed on the feature maps. 
While the striding increases the admissible motion magnitude of 
our cost volume, the local (sub-)pixel accuracy suffers.
Our solution is to refine the location 
by fitting a quadratic at the predicted motion and return its 
$\argmin$ and cost. 
To that end, we up-sample and scale the preliminary motion estimates 
$\bar{\mathbf{u}}$ and fit the quadratic in the vicinity of the 
initial match. Using nearest neighbor interpolation we can -- at each 
non-boundary pixel -- identify 4 motion candidates in the 
up-sampled motion field that are investigated for a refinement. 
Here, we fit a quadratic at the position (of the 4 candidates) 
of minimal cost and update the displacement to sub-pixel accuracy 
by solving the quadratic equation. 
%
To fit the quadratic, we utilize the high-resolution feature 
maps $\psi$, acquired before the striding in the feature generation 
takes place. 
We use a 5 stencil around the predicted match 
that is given by 
$(\bar{v}_0,\bar{v}_1):=\argmin_{v_0\in\{2\bar{u_0}, 2\bar{u_0}+1\},v_1\in\{2\bar{u_1}, 2\bar{u_1}+1\}} - \langle \psi^0(x,y), \psi^1(x+v_0,y+v_1) \rangle$.
We fit a quadratic function $f(\mathbf{v}):= \sum_i a_i v_i^2 + b_i v_i + c$ 
to the (negative) correlation 
$q_{x,y,v_0,v_1} := -\langle \psi^0(x,y), \psi^1(x+\bar{v}_0+v_0,y+\bar{v}_1+v_1) \rangle$ 
between the respective feature vectors in both images. 
Abbreviating the negative correlation cost with $q_{x,y,v_0,v_1}$ 
the solution to the fitting problem becomes:
\begin{align}
a_0 &=\! \frac{q_{x,y,+1,0}\!+\!q_{x,y,-1,0} \!-\! 2q_{x,y,0,0} }{2}, \;\; b_0 = \frac{q_{x,y,+1,0} \! - \! q_{x,y,-1,0}}{2}, \;\; c=q_{x,y,0,0}
\\
a_1 &=\! \frac{q_{x,y,0,1}\!+\!q_{x,y,0,-1} \!-\! 2q_{x,y,0,0} }{2}, \;\; b_1 = \frac{q_{x,y,0,1} \!-\! q_{x,y,0,-1}}{2}\;
\textrm{ and solution }
\\
v_0 &= \frac{-b_0}{2 a_0}, \quad v_1 = \frac{-b_1}{2 a_1} \quad \textrm{ with }
f(v_0,v_1) = a_0 v_0^2 + b_0 v_0 + c + a_1 v_1^2 + b_1 v_1.\label{eq:quadfit2}
\end{align}
%
If the fitting fails, 
the motion estimate $\bar{\mathbf{v}}$ is not a local minimum of the cost \wrt 
%
$\psi$ and we return the integral flow $\bar{\mathbf{v}}$ and 
cost at that pixel. 
Otherwise we set $\hat{\mathbf{u}}:= \bar{\mathbf{v}} + \mathbf{v}$.
The second row of \Fig\ref{fig:SintelDemo} shows a typical result 
for $\hat{\mathbf{u}}$ of our fitting layer. 
Subpixel accuracy can only be achieved if the initial discrete flow 
$\bar{\mathbf{u}}$ returned as the $\argmin$ solution of the correlation
volume is close to the ground-truth. 
We embed our fitting procedure into a single network layer. 
Backpropagation only requires us to compute the derivatives of the 
expression above \wrt the feature vectors $\psi$. 
Empirically, compared to directly matching features at the finest resolution, 
our combination of striding and quad-fitting leads to a smaller sized 
correlation volume and sub-pixel accuracy. 
%
%

\subsection{Diffusion Tensor}\label{sec:diff}
To compute the diffusion tensor, we apply a simple 4 layer 
feed-forward network with $64$ channels and relu non-linearities 
between $3\!\times\!3$ convolutions with dilations of 
size $1,2$ and $4$ on the input image $I^0$. 
%
By using dilations we operate at the finest resolution, 
but aggregate information of a larger spatial context \cite{YuKoltun2016}. 
%
%
To obtain values between 0 and 1 for our 2D tensor $W$, 
a final $1\!\times\!1$ convolution is followed by a sigmoid activation function. 
\Fig~\ref{fig:SintelDemo} ($3^\textrm{rd}$ row) shows a typical tensor 
obtained from an image of the Sintel benchmark. 
The regularization force is only low near image edges. 
Otherwise the regularizer of \eqref{eq:func_tv} operates at full strength.
%

\subsection{Confidence estimate}\label{sec:confidence}
Again we prefer a simple network to produce a single confidence 
estimate per pixel. 
The input to this stage are the upsampled pseudo-likelihoods \eqref{eq:likelihood}
of the preliminary motion estimates $\bar{\mathbf{u}}$, 
the upsampled distances between forward and backward flow vectors, $d(\bar{\mathbf{u}}, \bar{\mathbf{u}}^\textrm{BW})$, 
the distance to the boundary of the warped pixels, $b(\hat{\mathbf{u}})$,
and the costs after the quadratic fit \eqref{eq:quadfit2}. 
The forward-backward distance is computed by linear 
interpolation of the backward flow vectors, \ie 
$d(\bar{\mathbf{u}}, \bar{\mathbf{u}}^\textrm{BW}):=|\bar{\mathbf{u}}(x,y) - \bar{\mathbf{u}}^\textrm{BW}(x+\bar{u}_0,y+\bar{u}_1)|$. 
%
%
Assuming images of $N\!\times\! M$ pixels
the distance to the boundary is given by
$b(\hat{\mathbf{u}}):= \max(0,\min(x+\hat{u}_0,y+\hat{u}_1,N-x-\hat{u}_0,M-y-\hat{u}_1))$.
We further apply non-minimum suppression on the high 
resolution confidence map. 
Of the four pixels covered by one pixel of 
the low dimensional feature map, 
we only allow $c>0$ for the one of maximum confidence.
%

We posit, that all inputs to the network contribute some independent 
information that can be exploited for our confidence score:
a forward-backward check can identify occlusions and demand 
consistency, the $\argmin$ flow can lead to higher confidence 
in smooth regions than in noisy ones, 
a criterion also investigated in \eg~\cite{Hu2012}. 
By design, the softmax-probabilities are directly related to the confidence, 
whereas the fitting score can provide similar information, 
but in a more local vicinity. 
%
The network architecture is again given by a sequence of dilated 
$3\!\times\!3$ convolutions with relu activations, followed by a 
single $1\!\times\!1$ convolution with a sigmoid activation function. 
We employ seven layers with empirically chosen dilations of 
size $1,2,4,8,4,2$ and $1$ for the final $1\!\times\!1$ convolution 
that produces the 1D confidence signal. 
\Fig\ref{fig:SintelDemo} ($2^\textrm{nd}$ row, right) visualizes the 
confidence scores for the example. 
We observe that confident pixel are rare, and that those occur mostly close, 
but in some distance to the edges in the image. 



\subsection{Optimization Layer}\label{sec:opt}
Given the input from the stages above we can compute the solution to 
\eqref{eq:func_tv} or \eqref{eq:func_tgv} by unrolling iterations 
of a first-order algorithm. 
The non-smooth convex problem has a composite (smooth plus non-smooth)
structure that allows the application of FISTA \cite{Beck2009}, which
has optimal convergence rate $\frac{1}{k^2}$ after $k$
iterations \cite{Nesterov1983wy}.
Yet, the very sparse data term (\cf \Fig\ref{fig:SintelDemo}), 
leads to a situation that requires many iterations to achieve 
convergence for our diffusion like algorithm with local updates. 
A naive implementation in a high-level language like Tensorflow or 
Theano~\cite{TheanoShort16}, 
would demand $O(KNM)$ memory to perform backpropagation over $K$ iterations 
and $NM$ pixels, which is too much to handle for current GPU hardware. 
Likewise, accumulating gradients over many iterations in single precision 
can be numerically troublesome. 
Here, we show that it can be beneficial to implement the optimization 
stage of such first order algorithms, including backpropagation, 
within in a single layer in order to reduce memory overhead and 
make end-to-end training feasible. 
To that end, we combine two simple ideas, the gradient checkpointing technique 
\cite{Griewank2000} and accumulating the gradients into a buffer of 
double precision, such that the core of our algorithm can still run in the 
much more efficient single precision on the GPU. 
Checkpointing reduces the memory requirements to $O(\sqrt{K}NM)$ at the cost 
of a second forward pass. 
At $K\sim10^4$ this leads to savings of a factor of $100$ and, hence, 
we can backpropagate gradients over 10000 iterations without numerical problems.
%
In practice, we introduce $O(\sqrt{K})$ checkpoints every 
$\lceil{k \sqrt{K}}\rceil, k=0\ldots \lfloor\sqrt{K}\rfloor$ iterations 
at which we store the necessary information to reproduce a forward 
pass from $\lceil{k \sqrt{K}}\rceil$ to iteration $\lceil{(k+1) \sqrt{K}}\rceil$. 
This procedure demands $O(\sqrt{K}NM)$ storage. 
To perform backpropagation, we go backwards from checkpoint to checkpoint  
and recover the information needed to perform backpropagation 
between two checkpoints with a second forward pass. 
Again, this requires $O(\sqrt{K}NM)$ intermediate memory, which can be released afterwards. 
We then compute the gradients of this stage and, at the end, 
accumulate them into buffers of double precision. 

We focus our analysis on the TV like inpainting functional \eqref{eq:func_tv}. 
%
%
After discretization on a cartesian grid of size $N\!\times\! M$, 
the problem to determine $u_i\in \R^{NM}$, the pixelwise flow 
in each coordinate direction $i=0,1$ can be written as 
\begin{equation}\label{eq:tv_disc}
\min_{u_i} \|\sqrt{W} D u_i\|_\delta + \|u_i-\hat{u}_i\|_c. 
\end{equation}
The linear mapping $D\!:\!\R^{NM}\!\rightarrow\!\R^{2NM}$ 
approximates the spatial gradient of the flow per direction
via finite forward differences and $W\!:\!\R^{2NM}\!\rightarrow\!\R^{2NM}$ 
represents the discretized diffusion tensor and accordingly weighs 
the contribution of each local gradient per direction. 
The Huber norm $\|\cdot\|_\delta$ operates on each local gradient individually. The norm $\|\cdot\|_c$ denotes the $\ell_1$ norm weighted by the confidence $c$.
Our optimization procedure can be described by the following FISTA step:
\begin{align}
    u_i^{k+0.5} &:= v_i^k - D\trans \frac{W}{\max( 1, |\sqrt{W}D v_i^k|_2/\delta ) } D v_i^k \label{eq:f_u05}
    \\
    u_i^{k+1} &:= \left\{
    \begin{array}{cl}
    u_i^{k+0.5} - c\; &\textrm{if } u_i^{k+0.5} - c > \hat{u}_i \\
    u_i^{k+0.5} + c\; &\textrm{if } u_i^{k+0.5} + c < \hat{u}_i \\
    \hat{u}_i\; & \textrm{else}
    \end{array}\right.\label{eq:f_u1}
    \\
    v_i^{k+1} &:= u_i^{k+1} + \frac{t^k-1}{t^{k+1}} (u_i^{k+1}-u_i^{k}), \label{eq:f_v1}
\end{align}
%
which is executed for $i\!=\!0,\!1$ in parallel and $K$ iterations. 
The solution returned, $u_i^{K}$, is one of the components of $\mathbf{u}$, 
and $t^k\in\mathbb{R}$ denote the step-sizes. 
In contrast to \cite{VogelP17} we leave learning 
the step-sizes for future work and use 
$t^{k}:=\frac{1+\sqrt{1+4t^{k-1}}}{2}$ for $k\!>\!0$ and $t^0\!:=\!1$, 
as advocated in the original FISTA paper. 
This scheme guarantees a $\frac{1}{k^2}$ convergence rate~\cite{Beck2009}. 
We can incorporate the Lipshitz constant $L$ into 
our Tensor $W\!:=\!1/L\!\cdot\! W$. 
Later, the gradients $\frac{\partial f}{\partial W}$ \wrt $W$ 
have to be adjusted accordingly, 
we simply return $L \frac{\partial f}{\partial W}$. 
Here, $L$ is the maximal eigenvalue of our system matrix $D\trans W D$, 
\eg for TV we have $L=8$.
%
Given the gradient of some loss function $f$ on our returned 
solution $u^{K}$ from the forward pass,
$\frac{\partial f}{\partial u^{K}}$, we have to 
compute the gradients \wrt our parameters $W, \mathbf{\hat{u}}$ and $c$ 
(and $\beta$ for TGV-inpainting).
Our backward algorithm is composed of the following steps and returns
$\frac{\partial f}{\partial \hat{u}_i}$, $\frac{\partial f}{\partial c}$, 
$\frac{\partial f}{\partial W}$ 
and $\frac{\partial f}{\partial u^0}$, the gradient \wrt the initial solution $u^0$. 
\begin{align}
  \frac{\partial f}{\partial \hat{u}_i} &:= \frac{\partial f}{\partial \hat{u}_i} + \frac{\partial f}{\partial u_i^{k+1}} \textrm{ if } c < |\hat{u}_i-u_i^{k+0.5} | \label{eq:duhat}
  \\
  \frac{\partial f}{\partial c} &:= \frac{\partial f}{\partial c} + \textrm{sign}(\hat{u}_i-u_i^{k+0.5}) \frac{\partial f}{\partial u_i^{k+1}} \textrm{ if } c \geq |\hat{u}_i-u_i^{k+0.5}| \label{eq:dc}
  \\
  \frac{\partial f}{\partial u_i^{k+0.5}} &:=
  \left\{
  \begin{array}{cl}
  \frac{\partial f}{\partial u_i^{k+1}}\; & \textrm{if } c \geq |\hat{u}_i-u_i^{k+0.5}|
  \\
  0 & \textrm{else}
  \end{array}\right. \label{eq:du05}
  \\
  \frac{\partial f}{\partial v_i^{k}} & := \left( I- \frac{\partial}{\partial v_i^{k}} \left( \frac{W}{\max( 1, |\sqrt{W}D v_i^k|_2/\delta ) } D v_i^k \right)\trans \right) D \frac{\partial f}{\partial u_i^{k+0.5}}\label{eq:update_dv}
  \\
  \frac{\partial f}{\partial W} & := \frac{\partial f}{\partial W} + \frac{\partial}{\partial W} \left(\frac{W}{\max( 1, |\sqrt{W}D v_i^k|_2/\delta ) } D v_i^k\right)\trans D \frac{\partial f}{\partial u_i^{k+0.5}}\label{eq:update_dw}
  \\
  \frac{\partial f}{\partial u_i^{k}} & :=  \frac{\partial f}{\partial u_i^{k}} + \left(1+\frac{t^{k-1}-1}{t^{k}}\right) \frac{\partial f}{\partial v_i^{k}}\label{eq:du_k}
  \\
  \frac{\partial f}{\partial u_i^{k-1}} & := -\frac{t^{k-1}-1}{t^{k}} \frac{\partial f}{\partial v_i^{k}}. \label{eq:du_minus1}
  \end{align}
Here, we use the outer products in (\ref{eq:update_dv}, \ref{eq:update_dw}) 
to achieve a compact notation. 
In our implementation, we exploit the extreme sparsity of the resulting matrix.

Most of the above lines are a direct application of the chain rule, however, 
we briefly show how to arrive at \eqref{eq:update_dw}. 
Looking at a single gradient of index $l$, 
$\frac{\partial f}{\partial W_l}$, 
the chain rule suggests that
%
$\frac{\partial f}{\partial W_l} = \sum_k \left(\frac{\partial f}{\partial u_i^{k+0.5}}\right)\trans \frac{\partial u_i^{k+0.5}}{\partial W_l}$  
and $\frac{\partial u_i^{k+0.5}}{\partial W_l} = D\trans \left( \frac{\partial}{\partial W_l} \frac{W}{\max( 1, |\sqrt{W}D v_i^k|_2/\delta ) } \right) D v_i^k$.
While this justifies our summation over the iterations, we also observe that 
$\frac{\partial}{\partial W_l} (\frac{W}{\max( 1, |\sqrt{W}D v_i^k|_2/\delta ) })$ 
is the zero matrix except at indices $l$ and $l+1$, where we assume \wlogg that 
$l+1$ indicates the co-dimension of the diffusion tensor at the pixel. 
If we let $W_l^{l+1}$ denote the $2\times 2$ submatrix at row and column indices $l,l+1$ 
of the diagonal matrix $\frac{W}{\max( 1, |\sqrt{W}D v_i^k|_2/\delta ) }$, 
we find 
$\frac{\partial f}{\partial W_l} \!=\! \left((\!D\frac{\partial f}{\partial u_i^{k+0.5}})_l^{l+1}\!\right)\trans
\!\!\left(\!\frac{\partial}{\partial W_l} W_l^{l+1} \!\right) \! \left(\!D v_i^k\!\right)_l^{l+1}$,
which equals $\frac{\partial f}{\partial W}$ of \eqref{eq:update_dw} at the respective index $l$. 
Finally, note that the matrix in \eqref{eq:update_dv} also has non-zero entries 
for -- in backward direction -- horizontal and vertical neighbors of a pixel. 

Algorithmically backpropagation works as follows. 
Our checkpoint variables are $u^k_i$ and $v^k_i$ from which we
recover all $u^{k+0.5}_i$ and $v^k_i$ for a whole stage. 
Given $\frac{\partial f}{\partial u^{K}}$, we once 
apply (\ref{eq:duhat}-\ref{eq:du05}) to initialize 
$\frac{\partial f}{\partial u^{K-0.5}}$ along with 
$\frac{\partial f}{\partial \hat{u}_i}$ 
and $\frac{\partial f}{\partial c}$. 
Per iteration $k$ we already know 
$\frac{\partial f}{\partial u^{k+0.5}}$, $u^{k-0.5}_i$ and $v^k_i$ 
and can execute \Eqs~\ref{eq:update_dv}--\ref{eq:du_minus1} followed by
\Eqs~\ref{eq:duhat}--\ref{eq:du05} for iteration number $k\!-\!1$ to 
recover $\frac{\partial f}{\partial u^{K-0.5}}$ and we can continue 
with the next iteration by repeating these steps. 
Because $\frac{\partial f}{\partial u_i^{k-1}}$ from \eqref{eq:du_minus1}
is required in \eqref{eq:du_k} the values have to be kept in memory 
for one iteration. 
%

%
The TGV case \eqref{eq:func_tgv} is slightly more involved, but 
the derivations are similar. 
%
Discretization leads to the objective: 
\begin{equation}\label{eq:tgv_disc}
\min_{u_i}\min_{w_{i}=(w_{i,0},w_{i,1}\!)\trans}
\|\sqrt{W} (D u_i - w_{i})\|_\delta +
\beta ( \|D w_{i,0}\|_\delta + \|D w_{i,1} \|_\delta ) +
\|u_i-\hat{u}_i\|_c, 
\end{equation}
with auxiliary variables $w_{i,j}\!\in\!\R^{NM}$, $i,j \in\{0,1\}$.
We define the operator $B:\R^{3NM}\rightarrow\R^{6NM}$ by stacking 
the linear operations in \eqref{eq:tgv_disc} into a single  
mapping and likewise define the matrices 
$V_\beta:\R^{6NM}\rightarrow\R^{6NM}$ and 
$V:\R^{6NM}\rightarrow\R^{6NM}$. 
With $I$ denoting the identity mapping, $I:\R^{4NM}\rightarrow\R^{4NM}$, 
we define $V_\beta$ by stacking the 
$W$ and $\beta I$ into a single diagonal matrix. 
%
To construct $V$ we omit the multiplication of the 
unit matrix $I$ with $\beta$. 
As before, the Lipshitz constant $L=\max(12,8\beta)$ 
is incorporated into the operators $V_\beta$ and $V$ and 
the respective gradient $\frac{\partial f}{\partial V_\beta}$ 
is adjusted; again we return $L \frac{\partial f}{\partial V_\beta}$. 
Then the forward path becomes:
  \begin{align}
      (u_i^{k+0.5},w_{i}^{k+1})\trans &:= (v_i^k,q_{i}^k)\trans - B\trans \frac{V_\beta}{\max( 1, |\sqrt{V}B (v_i^k,q_{i}^k)\trans|_2/\delta ) } B (v_i^k,q_{i}^k)\trans \label{eq:fg_u05}
      \\
      u_i^{k+1} &:= \left\{
      \begin{array}{cl}
      u_i^{k+0.5} - c\; &\textrm{if } u_i^{k+0.5} - c > \hat{u}_i \\
      u_i^{k+0.5} + c\; &\textrm{if } u_i^{k+0.5} + c < \hat{u}_i \\
      \hat{u}_i\; & \textrm{else}
      \end{array}\right.\label{eq:fg_u1}
      \\
      (v_{i}^{k+1},q_{i}^{k+1})\trans &:= (u_{i}^{k+1},w_{i}^{k+1})\trans + \frac{t^k-1}{t^{k+1}} (u_{i}^{k+1},w_{i}^{k+1})\trans - (u_{i}^{k},w_{i}^{k})\trans), \label{eq:fg_v1}
    \end{align}
The complete backward path for TGV inpainting 
is presented in the supplementary material. 
Algorithmically we operate in the same manner 
as described for the TV case. 
In both cases we set $\delta$ to $0.1$. 
Please note that instead of learning a single scalar $\beta$, 
the algorithm can be extended to learn a pixel-wise diffusion tensor 
that operates on the auxiliary variables $w_0,w_1$ in \eqref{eq:tgv_disc}. 

%
\myparagraph{Hierarchical Optimization. }
Although we do not experience problems when training our network 
for 10K iterations at full resolution, 
a simple hierarchical strategy proves to be more efficient. 
At first we solve \eqref{eq:tv_disc} at a lower resolution, using 
down-sampled versions of $c$, $W$ and $\hat{\mathbf{u}}$ to initialize $u^0$. 
%
At the next level $c$, $W$ and $\hat{\mathbf{u}}$ are set accordingly, 
but $u^0$ is initialized by up-sampling the solution of the 
coarser level. 
Requiring fewer iterations at the finest level, this strategy 
accelerates the optimization.
%
Here, we use 3 levels with $2$K, $2$K and $4$K iterations.
%

%
%

%

\subsection{Training}\label{sec:train}
%
We start with the feature generation and pretrain 
this part of the network. 
Apart from the log-likelihood term \eqref{eq:likelihood}, our loss function 
also considers the quad-fitting procedure. 
%
Given the ground-truth flow $\mathbf{u}^*$, we define the 
loss function 
\begin{equation}\label{eq:loss_data}
L_\textrm{cor}( \hat{u}_0,\hat{u}_1 ) := \sum_{(x,y)\in\Omega} \log p_0(x,y,u^*_0) + \log p_1(x,y,u^*_1) + \alpha\min(1,|\hat{\mathbf{u}}-\mathbf{u}^*|_\epsilon), 
\end{equation} 
where we use $\epsilon=0.01$ for the Huber-norm and define the lookup 
by rounding the continuous $\mathbf{u}^*$ and set $\alpha=0.1$. 
%
Further, we use the up-sampled softmax probabilities in \eqref{eq:loss_data}  
and take care to not pass the gradients implied by the quad-fit 
through to the $\argmin$ lookup and correlation volume. 
%
Equipped with a trained and, in this work from now, fixed feature 
generation part, we further train the rest of the network by 
measuring the Huber-norm \wrt the ground truth. 
We set our displacement range to $d=96$ in each direction and note 
that it is effectively doubled, due to our 
striding/quad-fitting methodology. 
All our trainings are run on full images without down-sampling or cropping the input.  

%

	\section{Evaluation}
	\label{sec:evaluation}
	%
%
\def\picsint{0.31572}
\begin{figure}[tb]
  \begin{center}
     \includegraphics[width=\picsint\linewidth]{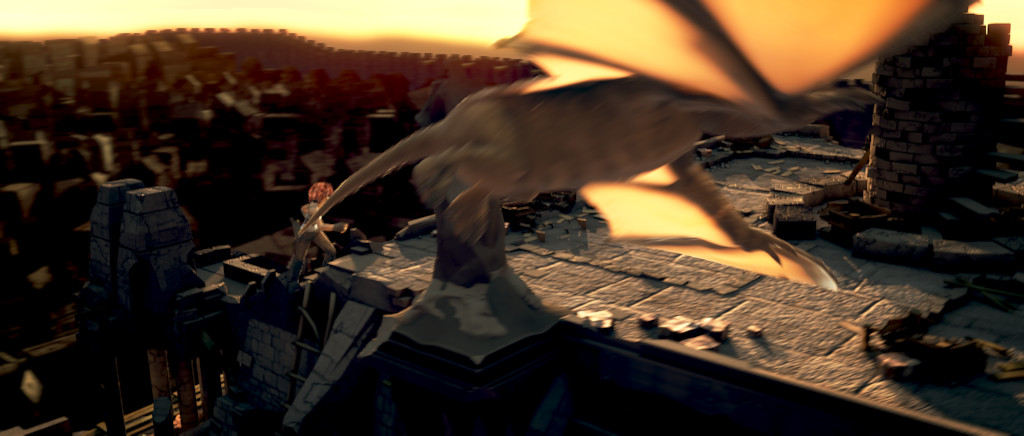}
     \includegraphics[width=\picsint\linewidth]{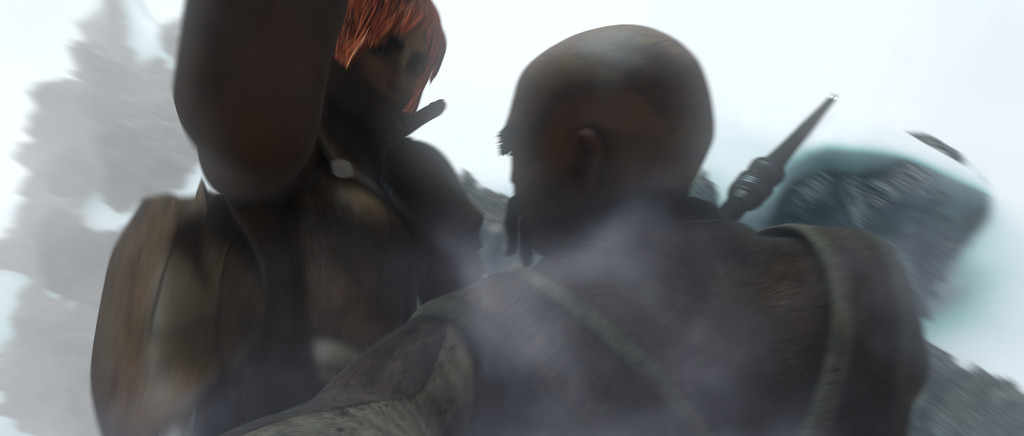}
     \includegraphics[width=\picsint\linewidth]{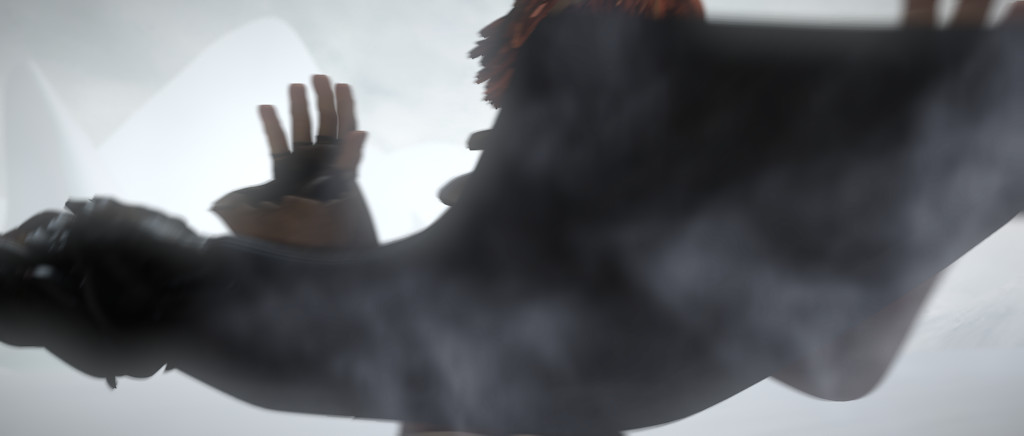}
     \\
     \includegraphics[width=\picsint\linewidth]{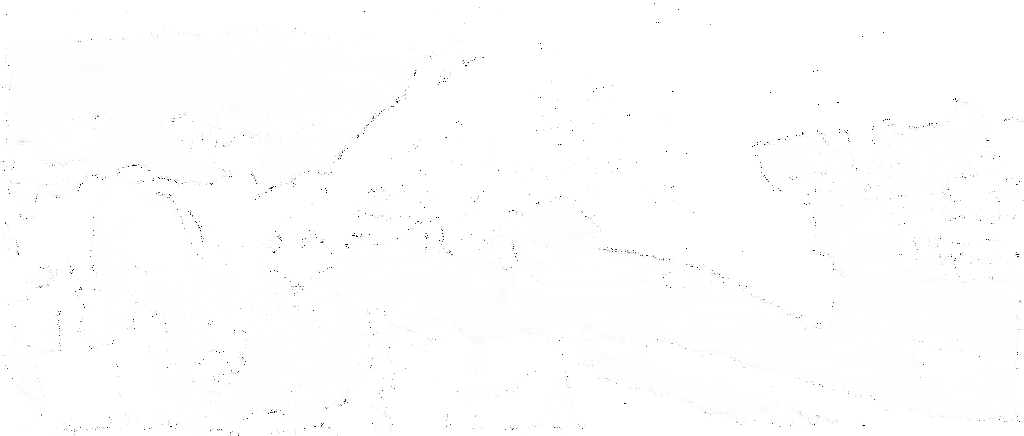}
     \includegraphics[width=\picsint\linewidth]{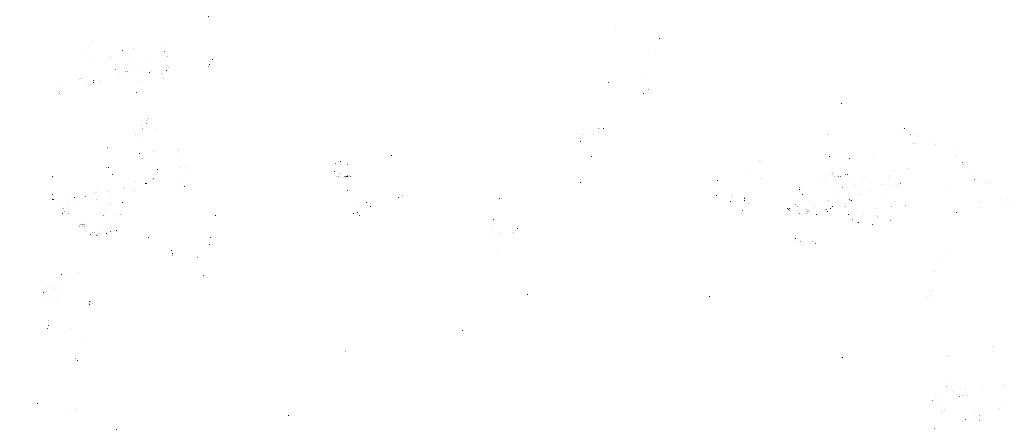}
     \includegraphics[width=\picsint\linewidth]{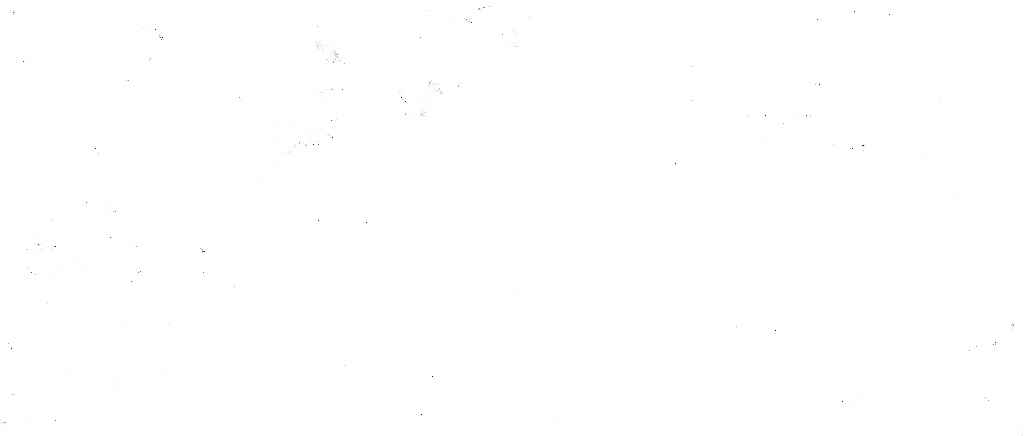}
     \\
     \includegraphics[width=\picsint\linewidth]{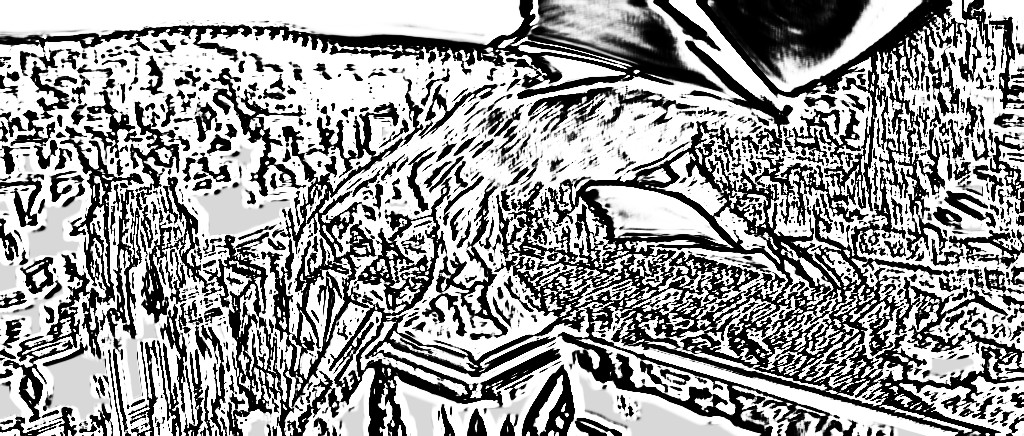}
     \includegraphics[width=\picsint\linewidth]{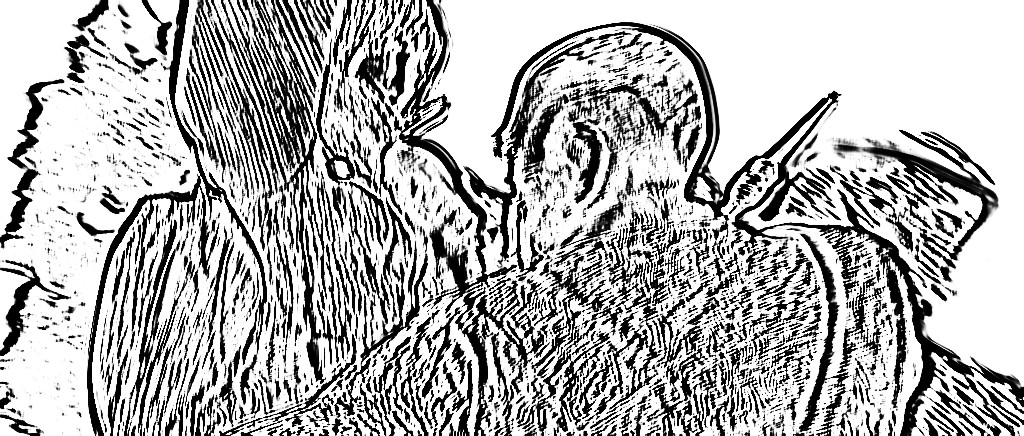}
     \includegraphics[width=\picsint\linewidth]{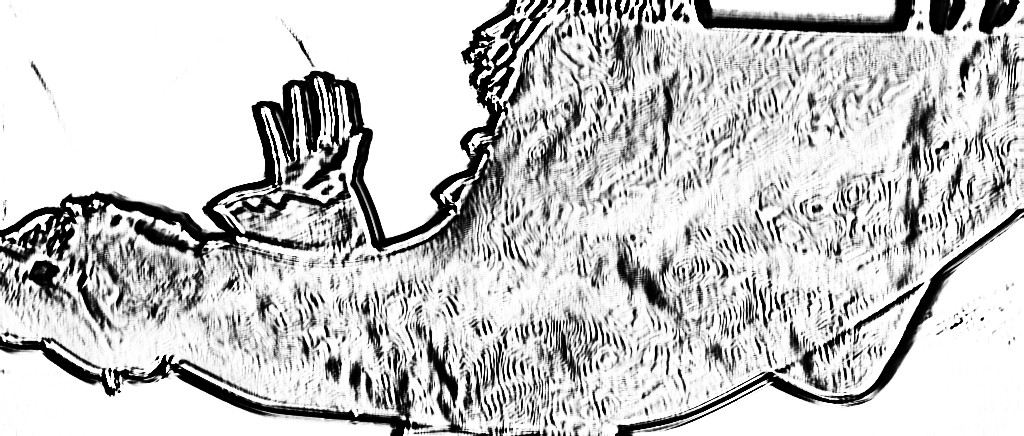}
     \\
     \includegraphics[width=\picsint\linewidth]{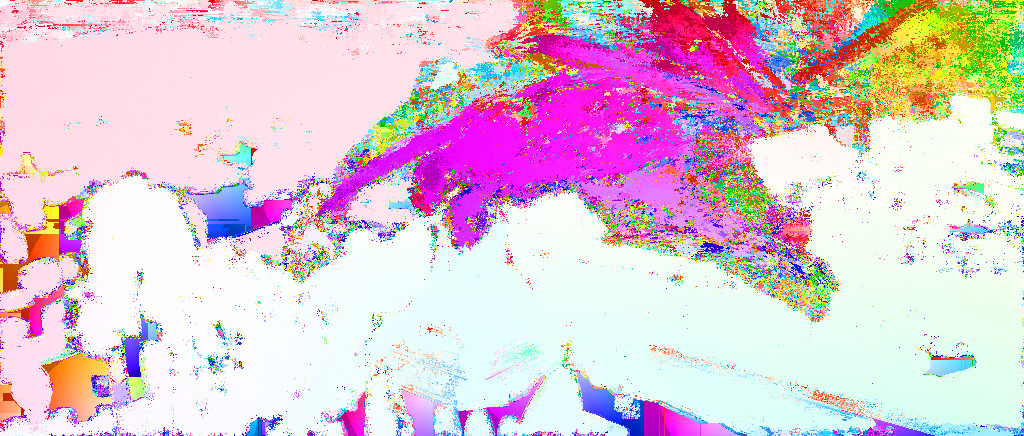}
     \includegraphics[width=\picsint\linewidth]{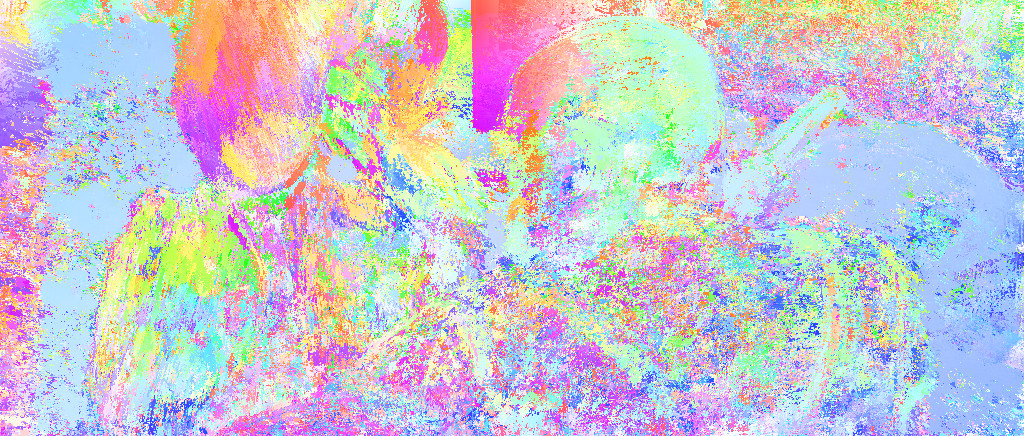}
     \includegraphics[width=\picsint\linewidth]{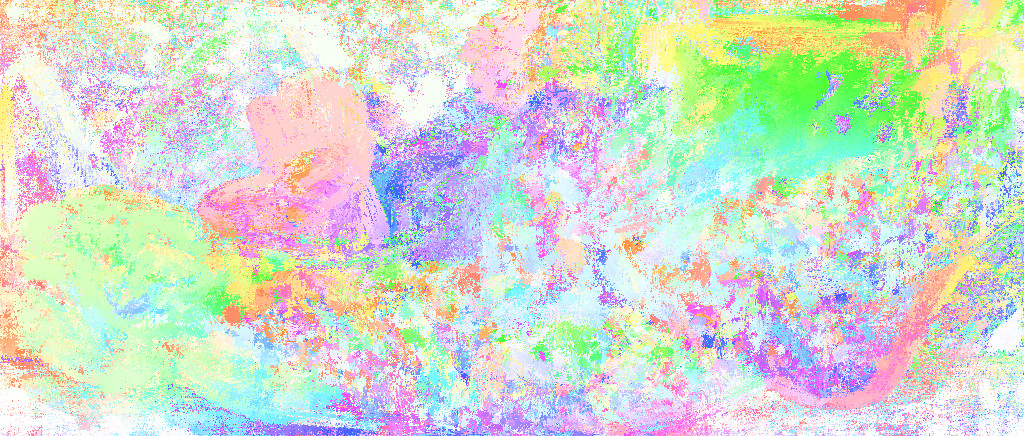}
     \\
     \includegraphics[width=\picsint\linewidth]{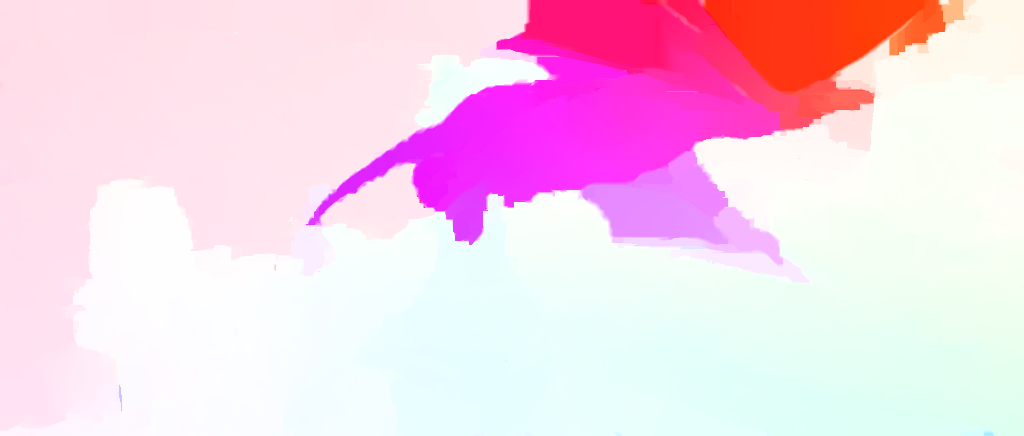}
     \includegraphics[width=\picsint\linewidth]{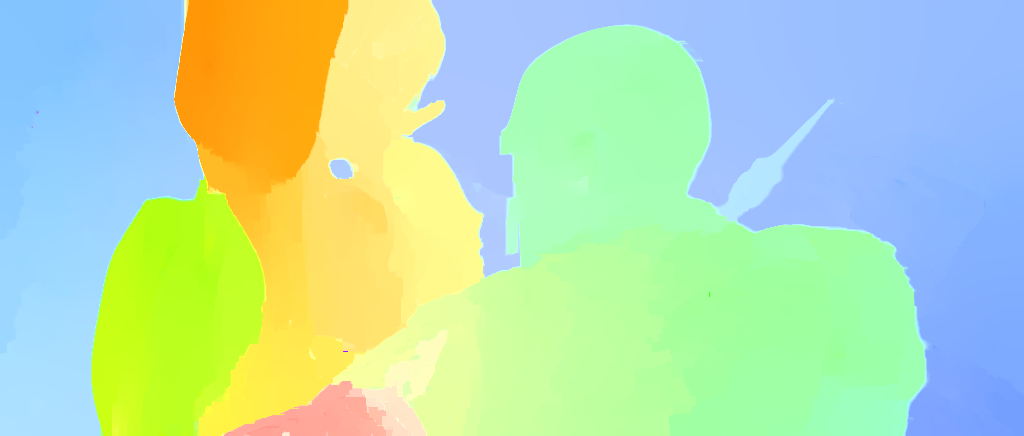}
     \includegraphics[width=\picsint\linewidth]{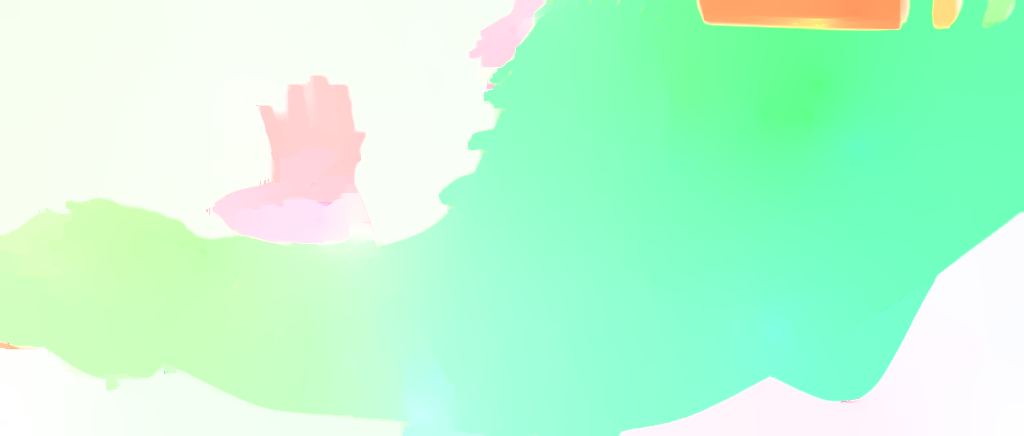}
     \\
     \includegraphics[width=\picsint\linewidth]{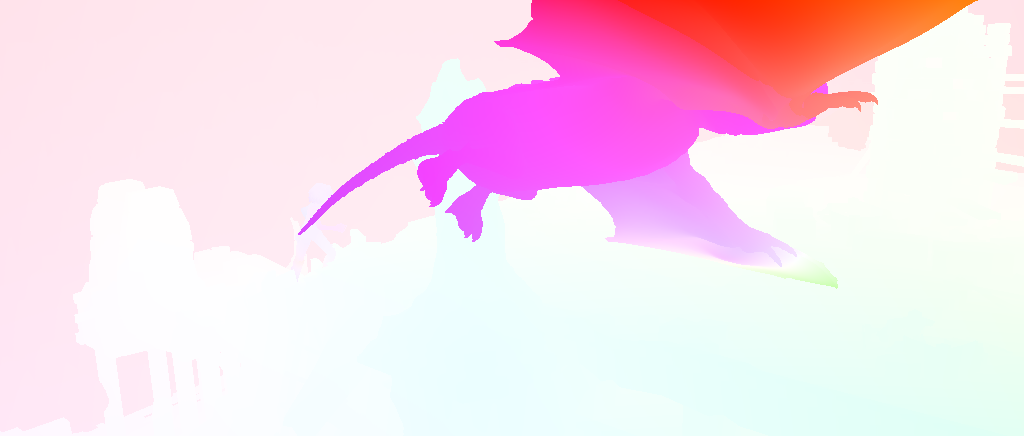}
     \includegraphics[width=\picsint\linewidth]{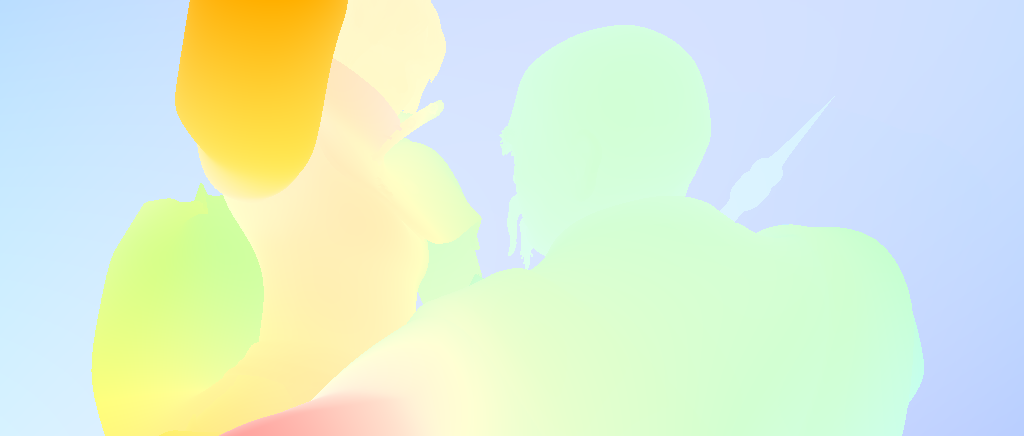}
     \includegraphics[width=\picsint\linewidth]{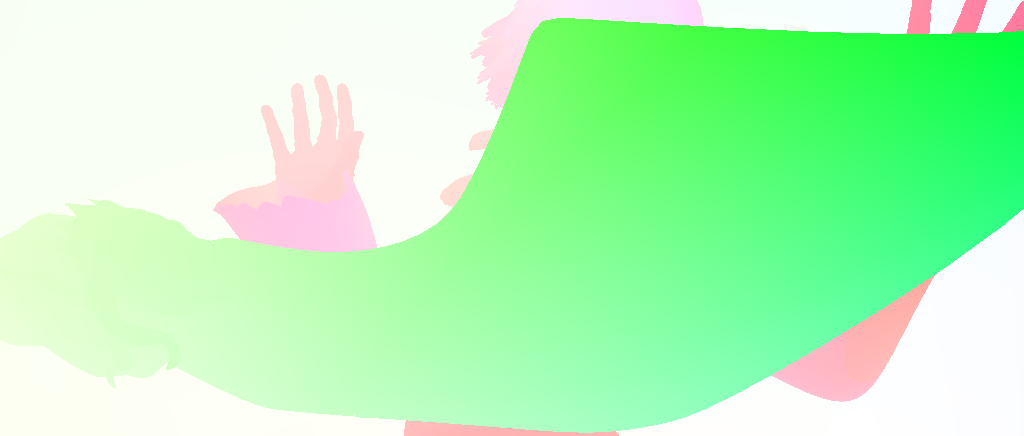}
  \end{center}
     \caption{
       Example results from the Sintel training data set~\cite{Butler2012}. 
       From top to bottom: input image $I^0$, confidence $c$,        
       horizontal component of the diffusion tensor $W$,
       $\argmin$ solution, solution after the optimization 
       layer and the ground truth.
     }\label{fig:Sintel}
\end{figure}
We train our networks with the Theano package~\cite{TheanoShort16}
on a machine equipped with a NVIDIA Titan X GPU with 12 GB of RAM. 
Our training data consists of either the artificial Sintel data set
\cite{Butler2012} or both of the KITTI~\cite{Geiger12,Menze2015CVPR} data sets. 
KITTI targets an automotive setting and provides real world scenes, 
but only sparse and approximately correct ground truth. 
When training on Sintel we use 441 of the 1041 images for training 
and use the other images for validation and evaluation. 
The KITTI data is divided in a similar manner. 
An implementation of our custom layers (quad-fitting, TV/TGV inpainting) 
can be found online\addtocounter{footnote}{-2}
\footnote{\url{https://github.com/vogechri/CustomNetworkLayers}}. 

\myparagraph{Memory and Runtime. } 
Overall our network has 450K parameters.
Training the feature generation on full sized Sintel images requires 4GB, 
the same amount is used for training the full inpainting network 
with backpropagation disabled for the pretrained part. 
%
In both cases, most of the memory is used by the correlation volumes, 
despite min-projection and down-sampling. 
%
A full joint training is left for future work. 
Our optimization layer proves to be a lightweight contribution to the network, 
requiring only additional 600MB of GPU memory at training time. 
A forward pass of the network requires $0.4s$ on a full sized image. 
%

%
\def\picskit{0.30875} 
\begin{figure}[tb]
  \begin{center}
%
\includegraphics[width=\picskit\linewidth]{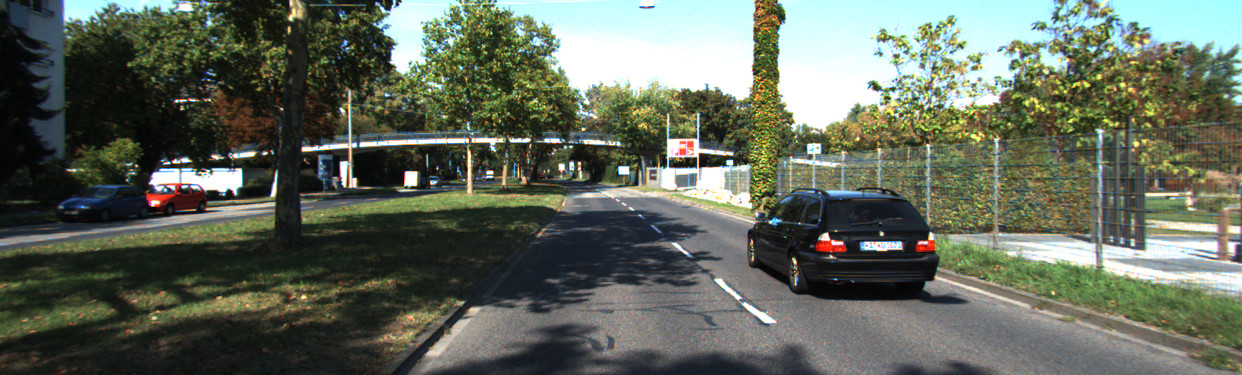}
\includegraphics[width=\picskit\linewidth]{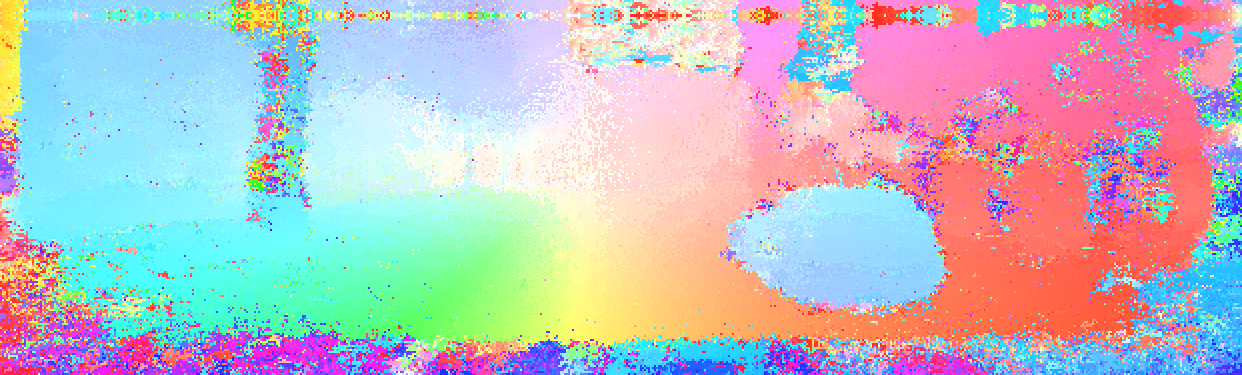}
\includegraphics[width=\picskit\linewidth]{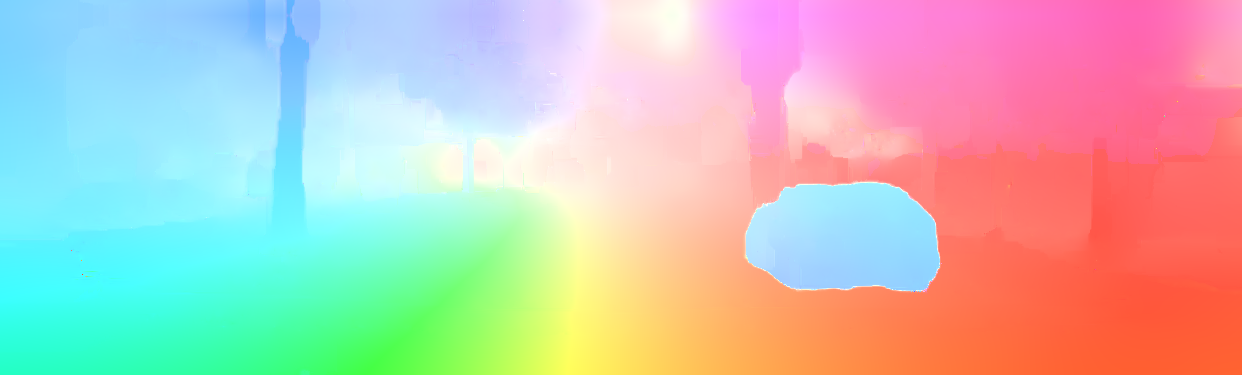}
\\
\includegraphics[width=\picskit\linewidth]{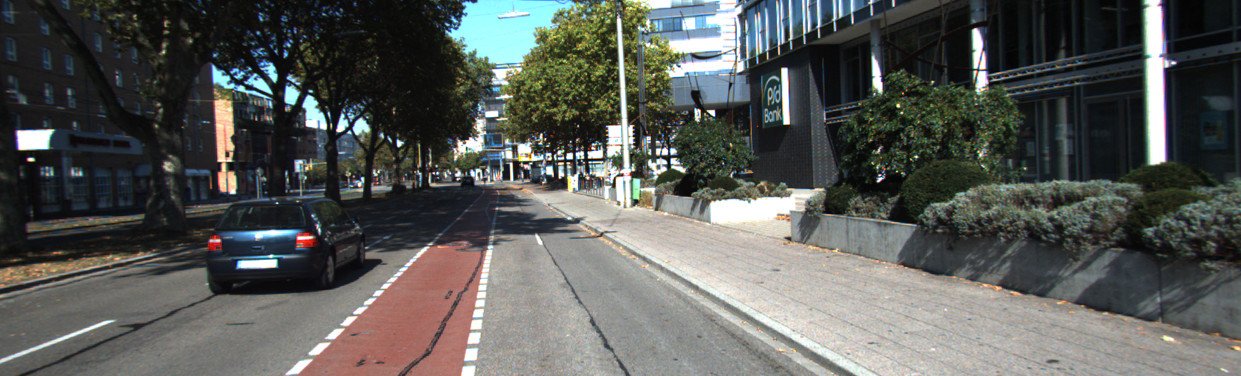}
\includegraphics[width=\picskit\linewidth]{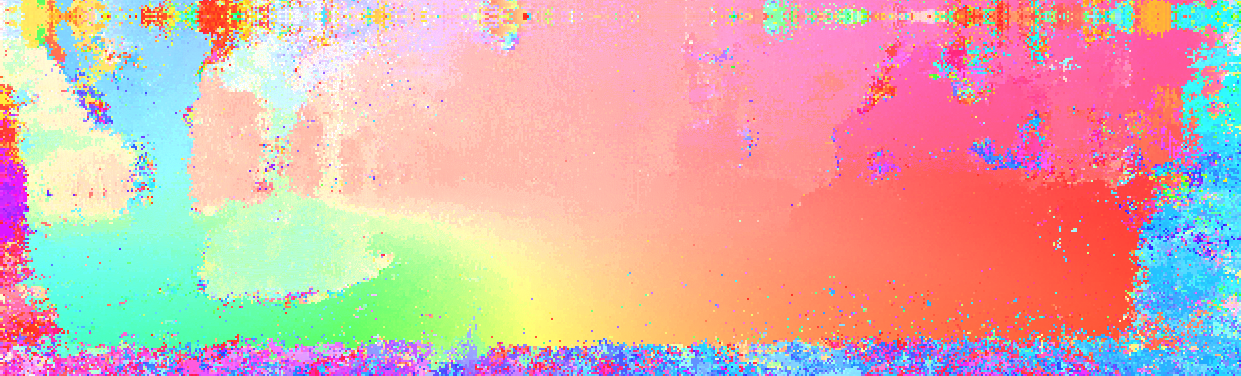}
\includegraphics[width=\picskit\linewidth]{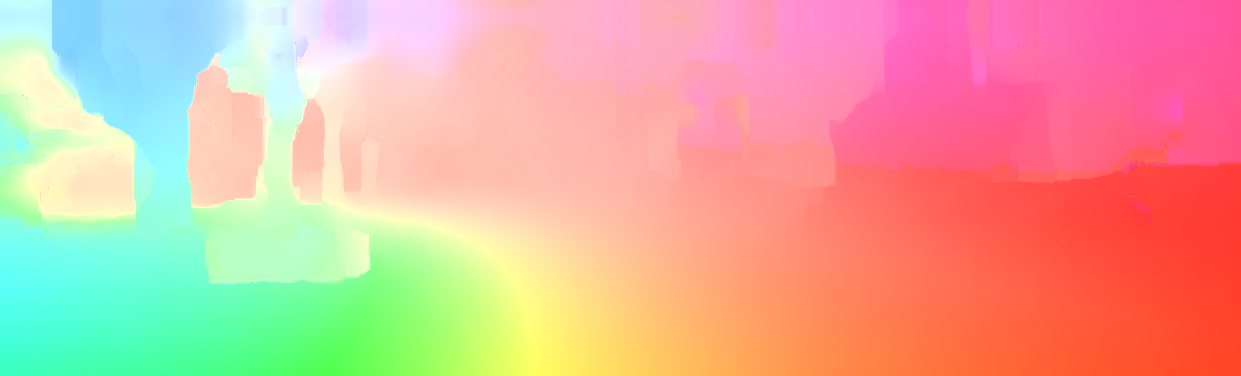}
\\
\includegraphics[width=\picskit\linewidth]{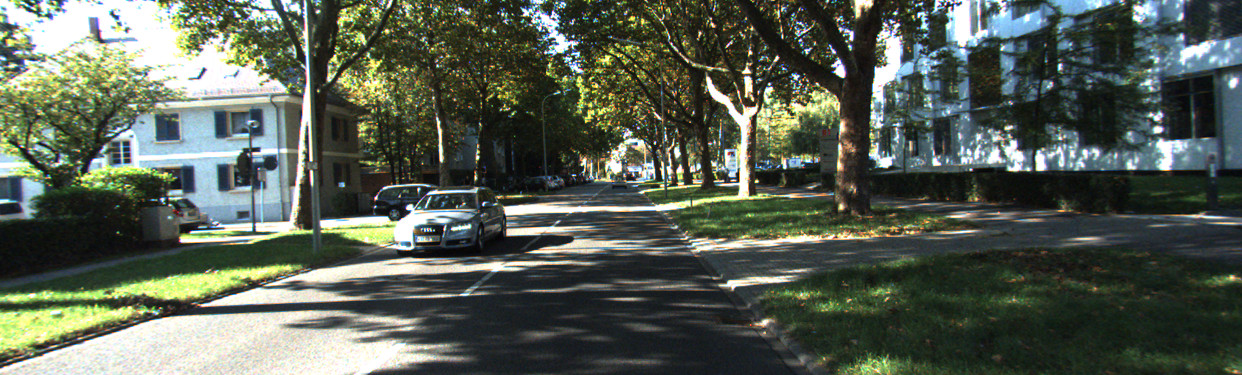}
\includegraphics[width=\picskit\linewidth]{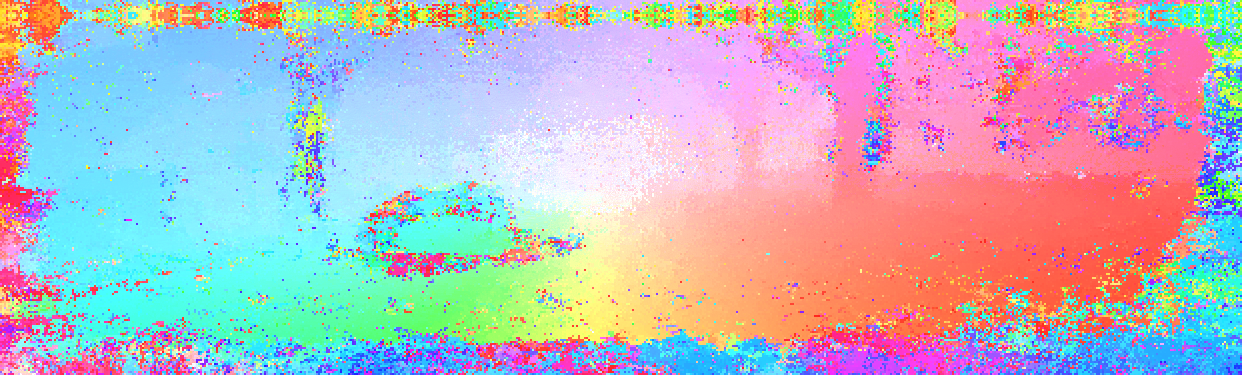}
\includegraphics[width=\picskit\linewidth]{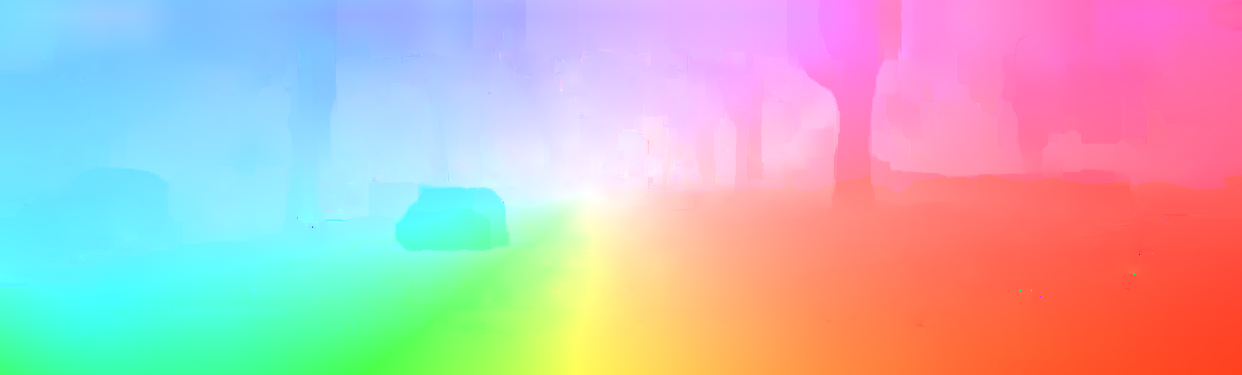}
  \end{center}
     \caption{
      \emph{Left}: Example from the KITTI training set~\cite{Menze2015CVPR}. 
     \emph{Middle}: The $\argmin$ solution, $\hat{\mathbf{u}}$, used as input to the optimization layer.
     \emph{Right}: Our inpainted solution (TGV variant).}
  \label{fig:Kitti}
\end{figure}
\subsection{Qualitative Evaluation}
%
We show a few results in \Fig\ref{fig:Sintel}
for the TV based inpainting on Sintel and in \Fig\ref{fig:Kitti} for our 
TGV model on KITTI. 
%
The hard Sintel examples lead to a noisy $\argmin$ solution 
($4^\textrm{th}$ row) and confident pixels ($2^\textrm{nd}$ row) are rare. 
The network appears to prefer only few correct matches, 
which in turn demands many iteration to spread the information. 
For the simpler left example the confident pixels outline the image 
and $\argmin$-flow edges (zoom in). 
Interestingly, compressive inpainting models~\cite{ChenXZG16} also 
prefer those kind supporting matches for their diffusion process. 
Those appear to possess the highest information content. 
%
%
%
The KITTI examples possess less sharp edges than the results 
on Sintel. 
We suspect the sparse ground truth, lacking information 
at object edges, to cause this phenomenon. 
\subsection{Quantitative Evaluation}
%
%
%
\begin{table}[!tb]
\setlength\tabcolsep{0.285cm} 
\centering
\newcolumntype{L}[1]{>{\raggedright\let\newline\\\arraybackslash\hspace{0pt}}m{#1}}
\newcolumntype{C}[1]{>{\center\let\newline\\\arraybackslash\hspace{0pt}}m{#1}}
\caption{Results on the Sintel training set (final). 
Displayed are end-point error (EPE) and percentage of 
outliers (Out), deviating $>\!3$ pixels from the ground truth 
for unoccluded (Noc) and all pixels (Occ).
Investigated methods include 
\emph{Quad-fit}: Results after the quadratic fit, 
\emph{TV-Model}:  TV inpainting~\eqref{eq:func_tv},
\emph{TGV-Model}: TGV inpainting~\eqref{eq:func_tgv},
\emph{$n\times$TV-Model}: TV inpainting trained using n times the default number of iterations in (\ref{eq:f_u05}-\ref{eq:f_v1}), 
\emph{$L_2^2$-Reg}: quadratic regularizer, 
\emph{$L_2^2$-All}: quadatric on confidence and regularizer.
} 
\label{tab:sintelTrain}
%
\begin{tabular}{L{2.4cm}cccc}
\rowcolor{gray!10}
Model & EPE (Noc) & $>3$px. (Noc) & EPE (Occ) & $>3$px. (Occ)  \\
\hline
Quad-fit     & 11.27 & 17.16\% & -- & -- \\ 
TV-Model     & 1.64  &  7.63\% & 2.48  & 10.29 \% \\ 
TGV-Model    & 1.81  &  7.79\% & 2.68  & 10.42 \% \\ 
$0.1\times$ TV-Model     & 3.11  &  10.01\% & 5.69  & 13.79 \% \\ 
$L_2^2$-Reg.(TV) & 2.21  &  8.38\% & 3.02  & 11.06 \% \\ 
$L_2^2$-All (TV) & 2.56  &  9.29\% & 3.70  & 12.23 \%  
%
%
%
%
%
%
%
%
%
\end{tabular}
%
\end{table}
%
%
\begin{table}[!tb]
\setlength\tabcolsep{0.15cm}
\centering
\newcolumntype{L}[1]{>{\raggedright\let\newline\\\arraybackslash\hspace{0pt}}m{#1}}
\caption{Results on the Kitti training set. 
Investigated methods include. 
\emph{TGV-Model}: TGV inpainting~\eqref{eq:func_tgv}.
\emph{bi-Laplacian}: Inpainting based bi-Laplacian described in the text. 
}
%
\begin{tabular}{L{2.2cm}cccc cccc}
\rowcolor{gray!10}
& \multicolumn{4}{c}{Noc} & \multicolumn{4}{c}{Occ}\\
\rowcolor{gray!10}
Model & EPE & 3px.  & 4px.  &5px. & EPE  & 3px.  & 4px. & 5px. \\
\hline

TGV-Model     & 1.73 & 8.16\%  &  6.33\%  &  5.26 \%  & 3.36  & 12.83\%  & 10.52\%  &  9.06 \% \\ 
TV-Model      & 1.85 &  9.90\% &  7.71\%  &  6.36 \%  & 6.93  & 23.53\%  & 19.43\%  &  16.53 \% \\
bi-Laplacian  & 2.19 & 10.42\% &  8.07\%  &  6.53 \%  & 4.17  & 15.70\%  & 12.97\%  &  11.38 \% \\
\end{tabular}
\label{tab:kittiTrain}
\end{table}
%
%
We start with the results for the Sintel data set in 
\Tab~\ref{tab:sintelTrain}. 
The TV-model can improve the 'matching only' solution after 
quad fitting by a significant amount. 
TGV performs about $10\%$ worse than TV. 
Note that for TGV more correct matches (3) are required to inpaint 
an affine motion in a segmented region than for TV (only 1). 
Training a TV-model with 10 times less iterations leads to 
significantly worse results. 
We observe that the higher the number of iterations, the sparser 
the confidence can be chosen; in other words, 
the stricter the selection becomes.
Finally, we also investigate if it is worthwhile to use a 
robust, non-smooth energy model or whether a linear model, 
as used in, \eg, \cite{BarronPoole2016} already suffices. 
In the fifth row we replace the Huber norm in \eqref{eq:func_tv} 
with a quadratic term. In the last row we introduce 
the square also on the $\ell_1$ term in \eqref{eq:func_tv}. 
The results gradually worsen the smoother the energy becomes. 
The complete linear model delivers results that are about 50\% worse. 
The network seems unable to compensate the lost robustness 
via diffusion tensor and confidence map alone. 

On KITTI (\Tab~\ref{tab:kittiTrain}) the TV based model cannot compete 
with the piecewise affine TGV model. 
We also compare inpainting with the robust TGV model \eqref{eq:func_tgv} 
with a well studied bi-Laplacian model: 
$\argmin_{u_i} \int_\Omega |W^\frac{1}{2} \Delta u_i|_2^2 + c|u_i-\hat{u_i}|_2^2 \dx$, 
$i=0,1$, where $\Delta$ is the Laplacian matrix. 
Again the linear model performs worse than our robust model, 
trailing its results by about $20\%$. 

\Tab~\ref{tab:allTest} compares our submitted models with a selection 
of inpainting based competitors on the official test set of both benchmarks. 
%
At first we notice a significant performance drop 
for both benchmarks, compared to the training set. 
%
On KITTI our network outperforms all three inpainting models 
Epic-Flow \cite{RevaudWHS15}, \cite{Hu2017CVPR} and the inpainting 
network of \cite{ZweigW17} by a large amount, although all either 
employ an affine model \cite{RevaudWHS15,Hu2017CVPR} 
or can learn one \cite{ZweigW17}. 
Our method even outperforms \cite{Guney2016ACCV}, 
who process the cost volume with a MRF model, before 
using \cite{RevaudWHS15} for inpainting. 
On Sintel our model performs on par or better than competing 
methods, but only for non occluded regions. 
Measuring all pixels our method trails 
\cite{Hu2017CVPR} and \cite{Guney2016ACCV}. 
However, recall that \cite{Hu2017CVPR} starts from a much better 
initial flow \cite{Hu2016CVPR} that is already on par with our 
approach in this metric. 
In fact many methods on the benchmark are not stand-alone, 
but utilize some well performing model for initialization 
or employ multiple post-processing steps. 
%
Here, our optimization layer could serve as a 
differentiable inpainting algorithm within a larger network. 

Our model trails the current state-of-the-art \cite{Sun_2018_CVPR,Hui_2018_CVPR} 
as shown \Tab~\ref{tab:allTest}. 
Yet, we use 10 (\cite{Hui_2018_CVPR}) or 17 (\cite{Sun_2018_CVPR})
times less parameters. 
Even \cite{spynet2017} uses more than twice the number of parameters
and clearly performs worse. 
Further, our model is minimalistic by design, employs no post-processing and 
can be used as complementary step for other methods, 
\ie by replacing our initial flow $\hat{\mathbf{u}}$ with theirs. 
%
%
%
\begin{table}[tb]
\setlength\tabcolsep{0.19cm} 
\centering
\newcolumntype{L}[1]{>{\raggedright\let\newline\\\arraybackslash\hspace{0pt}}m{#1}}
\newcolumntype{C}[1]{>{\center\let\newline\\\arraybackslash\hspace{0pt}}m{#1}}
%
\caption{Results on the Sintel~\cite{Butler2012} (final) and KITTI~\cite{Menze2015CVPR} test sets. 
Displayed are end-point error (EPE) for Sintel and percentage of outliers for KITTI.
Investigated methods include our submissions \emph{(TV)} and \emph{(TGV)}
and related methods from the benchmark. 
%
}
\begin{tabular}{L{2.649cm}cccccc|ccc}
  \rowcolor{gray!10}
  & (TV) & (TGV) & \cite{Hu2017CVPR} & \cite{ZweigW17} & \cite{RevaudWHS15} & \cite{Guney2016ACCV} &\cite{Hui_2018_CVPR} &\cite{Sun_2018_CVPR} &\cite{spynet2017} \\
  \hline
%
  \cite{Butler2012} EPE (Noc)              & 2.70 & -     & 2.77 & 2.79 & 3.06 & 2.62 & 2.4 & 2.4 & 4.5 \\ 
  \cite{Butler2012} EPE (Occ)              & 6.12 & -     & 5.62 & 6.04 & 6.29 & 5.73 & 5.4 & 5.0 & 8.4 \\
  \cite{Menze2015CVPR} $\!\%\!>\!3$px. (Noc) & -    & 10.8 & 10.3 & 13.6 & 16.7 & 12.4  & 5.5 & 5.1 & 26.7\\
  \cite{Menze2015CVPR} $\!\%\!>\!3$px. (Occ) & -    & 15.6 & 18.8 & 22.8 & 26.3 & 21.2  & 9.4 & 7.9 & 35.1
\end{tabular}
%
\label{tab:allTest}
\end{table}
	\section{Conclusion}
	\label{sec:conclusion}
We proposed a simple model for the accurate inpainting of optical flow, 
using an interpretable network to deliver the inputs 
for an optimization stage.
In the course of this work we proposed two non-custom layers, 
one for subpixel-refinement and one for optimization 
of our inpainting energy. 
For the latter, we showed that we can run and backpropagate 
through 10K iterations, which allows to accurately solve our 
energy problem. 
While the layer itself could be useful for various tasks, 
the lessons learned can also be transferred to similar 
problems. 

In the future we would like include 3D cost volume filtering 
and train the full model end-to-end, including feature generation. 
Further we believe that learning a selection mechanism 
for a joint TV/TGV regularization will be of benefit.

\bibliographystyle{splncs04}
\bibliography{bib}

\end{document}